\documentclass{article}

\PassOptionsToPackage{numbers, compress}{natbib}

\usepackage[final]{neurips_2022}

\usepackage[utf8]{inputenc} 
\usepackage[T1]{fontenc}    
\usepackage{hyperref}       
\usepackage{url}            
\usepackage{booktabs}       
\usepackage{amsfonts}       
\usepackage{nicefrac}       
\usepackage{microtype}      
\usepackage{xcolor}         
\usepackage{algorithm} 
\usepackage{algpseudocode}
\usepackage{graphicx}
\usepackage{amsmath}
\usepackage{subcaption}
\usepackage{comment}
\usepackage{tabularx}

\title{Dynamic Collaborative Multi-Agent Reinforcement Learning Communication for Autonomous Drone Reforestation}

\author{Philipp D. Siedler \\
Aleph Alpha\\
Stuttgart, Germany\\
\texttt{\{p.d.siedler\}@gmail.com}
}

\begin{document}

\maketitle

\begin{abstract}
We approach autonomous drone-based reforestation
with a collaborative multi-agent reinforcement learning (MARL) setup. Agents can communicate as part of a dynamically changing network. We explore collaboration and communication on the back of a high-impact problem. Forests are the main resource to control rising CO2 conditions. Unfortunately, the global forest volume is decreasing at an unprecedented rate. Many areas are too large and hard to traverse to plant new trees. To efficiently cover as much area as possible, here we propose a Graph Neural Network (GNN) based communication mechanism that enables collaboration. Agents can share location information on areas needing reforestation, which increases viewed area and planted tree count. We compare our proposed communication mechanism with a multi-agent baseline without the ability to communicate. Results show how communication enables collaboration and increases collective performance, planting precision and the risk-taking propensity of individual agents.

\textbf{Keywords}: Multi-Agent Reinforcement Learning; Graph Neural Network; Collaboration; Communication; Proximal Policy Optimization; Reforestation;
\end{abstract}
\begin{figure}[hb]
\begin{center}
\includegraphics[width=\textwidth]{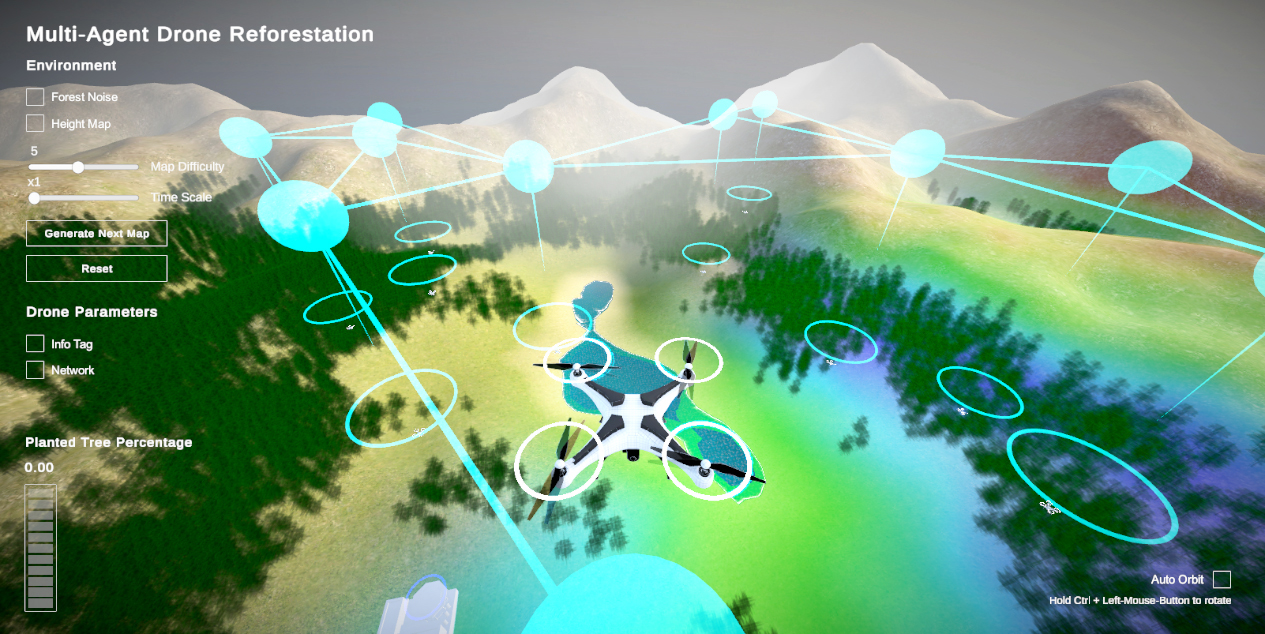}
\end{center}
\caption{Environment and heatmap indicating ideal areas for planting trees: green better, red worse. Web application: \url{https://philippds-pages.github.io/RL-Reforestation_WebApp/}.}
\end{figure}

\newpage

\section{Introduction}

\subsection{Motivation}



The success of our society is based on the human ability to communicate motives. Once intentions overlap amongst concerned parties, they may decide to join forces and collaborate to work towards a common goal that is beneficial to all and might not have been achieved by an individual. A group of people or individuals with specific interests can be seen as an agent, an entity with an agency. Most domains in the real world require collaboration to achieve higher goals and can be described as multi-agent (MA) systems.
Human intelligence is the highest observed, but we can draw further inspiration from nature. Species like bees only survive because of their ability to communicate and collaborate effectively. If a beehive successfully fends off predators, assigns tasks to collect nectar from flowers, and building materials, build and repair the hive, their survival is ensured \citep{bonabeau_swarm_1999}.\\
The last two years have brought climate change to the doorstep of many. Extreme heatwaves, wildfires and floods make life for animals and humans increasingly difficult. While vivid wildfires have destroyed large forest areas \citep{maccarthy_new_2022, tyukavina_global_2022}, cattle pasture is the biggest cause of deforestation \citep{dow_goldman_estimating_2020}. Forests absorb more than a quarter of global human-emitted carbon dioxide \citep{bernal_global_2018, tollefson_experiment_2013}. Droughts cause soil erosions and turn pasture and woodland into steppe. The Sahara desert advances irreversibly southward, causing livestock losses, human migration and mortality \citep{sendzimir_rebuilding_2011}. The Maradi and Zinder Regions have been regreening five million hectares to prevent further decline of the biosphere and habitability of the Niger \citep{pausata_greening_2020}. Multiple parameters can make reforestation difficult and must be considered to regreen successfully and sustainably. Trees need to be planted close to existing woodland to shield seedlings from harsh winds and provide them with stable soil. Furthermore, a mixture of native tree species needs to be planted rather than one kind. The additional maintenance, monitoring and transportation challenges, especially in hard-to-reach areas \citep{amigo_when_2020} have motivated this work. A wide variety of unresembling landscape scenarios makes this especially hard and requires coordination between multiple actors and information sources. \\
Core questions we ask: Can agents in a MA system learn the importance of communication and subsequently extend each other's partial observability? Furthermore, can agents learn to use the communicated information to increase individual performance? Can shared information be used to explore further and take actions with higher risks of depleting the battery for the benefit of finding locations with a higher need for reforestation and accordingly higher rewards? Finally, can agents organise themselves and delegate tasks, such that one group of agents become scouts and information collectors to extend the collectives' observable space and another group that carries tree seeds and plants them accordingly in areas they would not have found by themselves? Trying to answer these questions was the challenge of the experiments conducted and presented in this paper.

\subsection{Contribution}

In this work, we study communication in the context of a highly distributed, dynamic collective of autonomous drones and propose an approach to solve reforestation in hard-to-reach areas. Illegal logging has advanced to untouched areas of the Amazonas \citep{ennes_illegal_2021}. We propose to use autonomous drones to reach areas without infrastructure and roads, plant tree seeds and help monitor reforestation effectively. Controlling a single or collective of drones to fly hardcoded manoeuvres \citep{bogdanowicz_flying_2017} and in formation \citep{ahn_real-time_2019} is a solved problem. However, we are interested in solving the reforestation problem with a decentralised autonomous system consisting of individual agents capable of making ad-hoc decisions in various scenarios. In this work, we utilise Proximal Policy Optimisation (PPO) \citep{schulman_proximal_2017}, a state-of-the-art Reinforcement Learning algorithm to train a collective of multiple agents. The task for the MA collective is to pick up tree seeds, search for spots along the perimeter of an existing forest that benefit the most from reforestation and plant the carried tree seeds. Furthermore, the proposed agent collective can communicate to enable collaboration. Agents can capture spots they found in memory while exploring the environment and share the information with neighbouring agents when in reach. Each agent controls a drone and has to execute the planting tasks and search for new spots, but also has to monitor the battery charge status, especially with the additional payload when carying a tree seed. We have developed a simulation environment to train and test various learning mechanisms. Our environment is additionally able to cater for open-ended learning \citep{open_ended_learning_team_open-ended_2021} through an infinite procedural generated set of scenarios with possibly increasing difficulty of the terrain topology and forest sparsity. Our contribution is two-fold; we mainly want to demonstrate that a communication mechanism for a MARL system in a partially observable environment can lead to collaboration and increased collective performance amongst highly dynamic agents with a proximity-based communication network. Furthermore, as a secondary contribution, we want to do so by utilising an environment with a minimal gap to a high-impact real-world problem, namely autonomous drone-based reforestation in hard-to-reach areas \citep{kuttler_nethack_2020}.\\
We have designed a task to verify our solution. The task is constrained by a given time frame. Individual agents of the collective spawn at the drone station spawn point. Their initial state is a full battery charge and a loaded tree seed. They then have to search for a spot in urgent need of reforestation to collect a high reward. After agents have dropped their tree seed, they must return to the drone station before depleting their battery. This end-to-end task can be repeated until the episode ends. A MA setup with and without the ability to communicate is trained on a single scenario as well as on multiple scenarios, respectively. The setup without communication ability serves as the baseline. The benefit of communication can be verified by the count of trees planted, but more importantly, the quality and precision of the positioning of the planted tree seed regarding the existing forest volume. Finally, we demonstrate how well agents trained in multiple scenarios perform in an unseen scenario compared to those trained in a single scenario.

\subsection{Future Work}
The partial observability hinders the agents from keeping track of the big picture. For future work, we would be interested in studying how much of the overall forest in a scenario is explored and if we can observe that agents understand what areas of the forest need reforestation and find a tree seed distribution balance. Once an overall understanding is achieved, agents should be able to weight areas rather than planting tree seeds at the next available good spot greedily. This would require to increase the long-term thinking of agents. There are ideas to have rounds of exploration followed by regrouping and sharing of gathered information before tree seeds are carried and planted according to a plan that has been agreed on collaboratively. Furthermore, studying a larger information memory would be an exciting experiment strand. Questions we might ask are: What if agents can tell each other where a good planting spot is and where enough trees have been planted already, or if no good spots are to be found anymore? Finally, incentivise agents to allocate different tasks to each other, i.e. scout drones that can fly quicker with less energy that gather information, with higher flexibility to explore, and planter drones that pick up, carry and plant tree seeds accordingly.


\section{Related Work}

Single-Agent Reinforcement Learning (SARL) is a learning paradigm based on an agent taking actions in an environment to maximize cumulative reward. In contrast to other machine learning methods, RL does not require a dataset and learns from collected experiences. Nevertheless, just like a well-curated dataset is crucial for successful supervised learning, the experiences collected by an RL agent heavily depend on the environment design \citep{open_ended_learning_team_open-ended_2021}. There is broad interest for RL, from industry \cite{leitao_industrial_2015}, but also academia and research. Exemplary, industry applications include optimal control, autonomous vehicles \citep{shalev-shwartz_safe_2016}, robots \citep{kober_reinforcement_2013, sukthankar_cooperative_2017, ismail_survey_2018}, cooling management \citep{lazic_data_2018}, but also social dilemmas \citep{leibo_scalable_2021}, economics and finance \citep{charpentier_reinforcement_2021}. Many, if not all, real-world domains and according applications include multiple agents or at least some form of an entity with an agency, active or passive parts of the environment. However, experimenting in the real world can lead to high resource consumption and time spent. After all, an RL system does not only learn from succeeding in a given environment, but a failing experience can be crucial for a robust learning update. This constraint brings us to the need for simulation frameworks, where we can break things and even manipulate time. Furthermore, single- and multi-player games have been popular test-beds for algorithm and learning mechanism development. Board games like GO \cite{silver_mastering_2016,silver_mastering_2017}, Chess \citep{campbell_deep_2002}, Shogi \citep{silver_general_2018}, Hex \citep{anthony_thinking_2017}, Poker \citep{moravcik_deepstack_2017, brown_superhuman_2018} and Diplomacy \citep{anthony_learning_2022, calhamer_diplomacy_1959}, but also computer games i.e. Atari \citep{mnih_human-level_2015}, Dota \citep{openai_dota_2019}, Starcraft \citep{vinyals_grandmaster_2019} and overcooked \citep{fontaine_importance_2021} have shaped the RL field significantly. The combination of simulation and the game domain naturally leads us to game engines, including realistic physics \citep{ward_using_2020}, which we will be using for our work.\\
Collaboration \cite{cohen_team_1997,decker_distributed_1987,pynadath_communicative_2002} in MA systems can be achieved in various ways and has a rich history of literature \citep{shoham_multiagent_2009}. Collaboration does not need active communication \citep{matignon_independent_2012, panait_cooperative_2005} and can be achieved through methods such as gradient-based distributed policy search \citep{peshkin_learning_2000}, reward function sharing \citep{lauer_algorithm_2000}, memory sharing \citep{lowe_multi-agent_2017, pesce_improving_2020, hernandez-leal_survey_2019} and parameter sharing (PS) \citep{sukthankar_cooperative_2017, hernandez-leal_survey_2019}. Nevertheless, our work focuses on active communication as part of the agents' action space. \citep{xuan_communication_2001}. The messages each agent can communicate consist of a three-dimensional vector and represents a location in euclidean space \citep{mataric_using_1998}. The agent can decide to save a location as a three-dimensional vector that is worth memorizing and send it to the three nearest neighbouring agents in reach. In contrast, other work proposes communicating more complex information such as intentions or policy gradients \citep{foerster_learning_2016}. Our communication layer is based on Graph Neural Network (GNN) \citep{scarselli_graph_2009} message passing \citep{gilmer_neural_2017}, as part of the agents observation space. This enables the agent to deal with various sized graph-structured data. Recent work of ours has investigated highly decentralized multi-agent communication, with fixed size graph-structured data \citep{siedler_collaborative_2022}, as well as variable size graph-structured data \citep{siedler_power_2021}, with communicating as part of the action space. While there has been work on static networks \citep{haksar_distributed_2018}, in contrast to our previous work, here we advance static- and investigate dynamically-changing graph-structured data with a maximum size \citep{zhang_efficient_2019, ding_learning_2021}.\\
There have been multiple single and multi-drone applications in various fields. Previous work has investigated the use of drone networks \citep{julian_image-based_2019} to fight forest fires by monitoring outbreaks and growth \citep{afghah_wildfire_2019}. Furthermore there is work on last-mile delivery based on single autonomous drones \citep{campbell_will_2022, sorooshian_toward_2022, reese_teamsters_2022}. We think it is important to mention two companies, namely Flash Forest \citep{flash_forest_flash_nodate} and Airseed \citep{airseed_airseed_nodate}, using Artificial Intelligence (AI) to solve reforestation with drones. Both companies, but also academic work \citep{lohit_reforestation_2021} heavily use mapping, pre-planning of flight paths and tree seed drop locations. However, our proposal investigates fully autonomous flight-path planning, drop location scouting and communication amongst all agents, re-charging and battery-life tracking using a single neural network for each drone. Furthermore, we are not trying to replace the proportional integral derivative (PID) controller. The PID is commonly used to make the drone pitch, yaw and roll \citep{labayrade_single_2003}. While there is work using a neural network to replace the PID and fully control the drone's movement \citep{venkatesh_fully_2017}, this is out of scope for this work. Instead, we are using a simplified set of commands, actuating the drone to move.

\section{Background}


\subsection{Proximal Policy Optimisation (PPO-Clip)}


Proximity Policy Optimisation (PPO), a state-of-the-art, on-policy RL algorithm, has been utilised for training the agents in this work. Our environment's action space consists of discrete and continuous actions, which PPO supports. Two main concepts define the PPO algorithm: 1. PPO performs the largest possible but safe gradient ascent learning step by estimating a trust region, and 2. Advantage estimates how good an action in a specific state is, compared to the average action. Various other RL algorithms, such as Asynchronous Advantage Actor Critic (A3C), use this concept \citep{udacity-deeprl_introduction_2019}.
\textbf{Advantage:} Advantage can also be described as the difference of the Q Function and the Value Function: $A(s,a) = Q(s,a) - V(s)$, where $s$ is the state and $a$ the action \citep{zychlinski_complete_2019}.
The Q Value (Q Function), denoted as $Q(s,a)$,
measures the overall expected reward given state $s$, performing action $a$. Assuming the agent continues playing until the end of the episode following policy $\pi$. The Q is abbreviated from the word Quality, and denoted as: $\mathcal{Q}(s,a) = \mathbb{E}\left[ \sum_{n=0}^{N} \gamma^n r_n \right]$.
The State Value Function, denoted as $V(s)$, measures, similar to the Q Function, overall expected reward, with the difference that the State Value is calculated after the action has been taken and is denoted as: $\mathcal{V}(s) = \mathbb{E}\left[ \sum_{n=0}^{N} \gamma^n r_n \right]$. The Q Value $V(s)$, with $n=0$, is the expected reward $r^0$ in state $s$, before action $a$ was taken, while the Q Value measures the expected reward $r^0$ after $a$ was taken.
\textbf{Trust Region:} After some experiences $\pi_{\theta_k}(a_t|s_t)$ have been collected, the trust region can be calculated as the quotient of the current policy to be refined $\pi_\theta(a_t|s_t)$ and the previous policy as follows $r_t(\theta) = \frac{\pi_\theta(a_t|s_t)}{\pi_{\theta_k}(a_t|s_t)} = \frac{current\ policy}{old \  policy}$.
This is a simplified gradient ascent objective function with limited deviation between the current and old policies \citep{achiam_simplified_2018}.
\begin{quote}
$\mathcal{L}_{\theta_k}^{CLIP}(\theta) = \underset{s,a\sim\theta_k}{\mathbb{E}} \left[\min{\left( r_t(\theta)A^{\theta_k}(s,a), g(\epsilon,A^{\theta_k}(s,a))\right)}\right]$,
\newline
\newline
where
\newline
\newline
$g(\epsilon,A) = 
\begin{cases}
(1 + \epsilon)A,& \text{if } A\geq 0\\
(1 - \epsilon)A,& \text{otherwise}
\end{cases}
$
\end{quote}
The advantage function will be clipped if the probability ratio between the current and the previous policy is outside the range of ($1+\epsilon$) and ($1-\epsilon$). This also means that the advantage will never exceed the clipped values and prevents the new policy from getting too far from the old policy. In the original PPO paper by Schulman, et al. (2017) \citep{schulman_proximal_2017} $\epsilon$ was set to 0.2.
Lastly, the policy that yields the highest sum over all Advantage estimates $A_t$ in range of max time step $T$ of a trajectory $\tau \in \mathbb{D}_k$ will be used to override the old policy $\theta_{old}$ \citep{openai_proximal_2021}: $\theta_{k+1} = arg \underset{\theta_k}{max}\frac{1}{|\mathbb{D}_k|T}\sum_{\tau \in \mathbb{D}_k}\sum_{t=0}^{T}\min\left( \frac{\pi_\theta(a_t|s_t)}{\pi_{\theta_k}(a_t|s_t)},g(\epsilon,A^{\theta_k}(s,a)) \right)$.

\subsection{Graph Neural Network}

\begin{figure}[H]
    \begin{subfigure}{0.4\textwidth}
        \centering
        \includegraphics[width=0.6\linewidth]{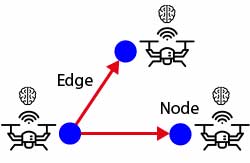}
        \caption{Directed Graph}
        \label{fig:directed-graph}
    \end{subfigure}
    \begin{subfigure}{0.6\textwidth}
        \centering
        \includegraphics[width=0.8\linewidth]{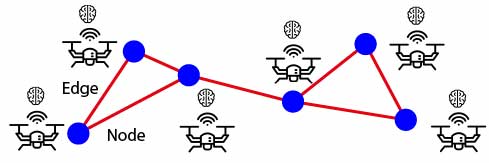}
        \caption{Undirected Graph}
        \label{fig:undirected-graph}
    \end{subfigure}
    \caption{Graph $\mathcal{G}$ consisting of vertices $\mathcal{V}$ (blue dots) and edges $\mathcal{E}$ (red lines): $\mathcal{G} = (\mathcal{V},\mathcal{E})$}
\end{figure}

Scarselli, et al. (2009), developed fundamental work on Graph Neural Networks, leading to many variations, such as Gated Graph Sequence Neural Networks \citep{li_gated_2017}, Graph Attention Networks \citep{velickovic_graph_2018} and Convolutional Neural Networks on Graphs with Fast Localized Spectral Filtering \citep{defferrard_convolutional_2017}.
A graph is a data structure based on nodes or vertices and edges. The node and edge objects can hold an arbitrary amount of features of any type. An edge represents the relationship between two nodes, but a node can have unlimited relationship edges with other nodes. Edges can be directed from node A to B \ref{fig:directed-graph}, or undirected, from node A to B and vice versa \ref{fig:undirected-graph}.

Fundamental functionalities of GNNs are graph, node and edge classification. Features and the existence of nodes and edges can be predicted using, i.e. neighbouring nodes and existing edges. Classification of a graph as a whole can be achieved using node and edge features and the graph's topology as input. However, one of the simplest forms of a GNN is the message passing framework proposed by Gilmer, et al. \citep{gilmer_neural_2017}, using the "graph-in, graph-out" network architecture introduced by Battaglia, et al. \citep{battaglia_relational_2018}. Hence the graph topology is not modified but loaded node and edge feature embeddings.

Node states can be denoted as $v$, edges connected to node $v$ as $x_{co[v]}$. The state of a node $h_v$ may consist of a n-dimensional vector feature. Adjacencies between a node and its neighbours are the mapped transition of the node, denoted as $h_{ne[v]}$, including all neighbouring node features, denoted as $x_{ne[v]}$. The transition function $f$ is used to embed each node on a n-dimensional space \citep{zhou_graph_2020}: $h_v = f(x_v, x_{co[v]}, h_{ne[v], x_{ne[v]}})$ \\
Breadth-First Search (BFS) \citep{burkhardt_optimal_2021}, Depth-First Search (DFS) \citep{kaur_analysis_2012} and random walk based DeepWalk \citep{perozzi_deepwalk_2014} are popular algorithms to define neighbourhoods in graph structured data, however we define neighbourhoods using $n$ nearest neighbours based on shortest euclidean distances.
Furthermore, passing state $h_v$ and feature $x_v$ of nodes and edges, to the GNN outputs the result of function g: $o_v = g(h_v, x_v)$.
And finally, applying gradient descent to formulate loss using the ground truth $t_v$ as well as the output $o_v$ of node $v$: $loss = \sum_{i=1}^{p}(t_i - o_i)$.

\section{Method}

\subsection{Drone Reforestation Environment}

\begin{figure}[!ht]
\begin{subfigure}{0.19\textwidth}
\centering
\includegraphics[width=\textwidth]{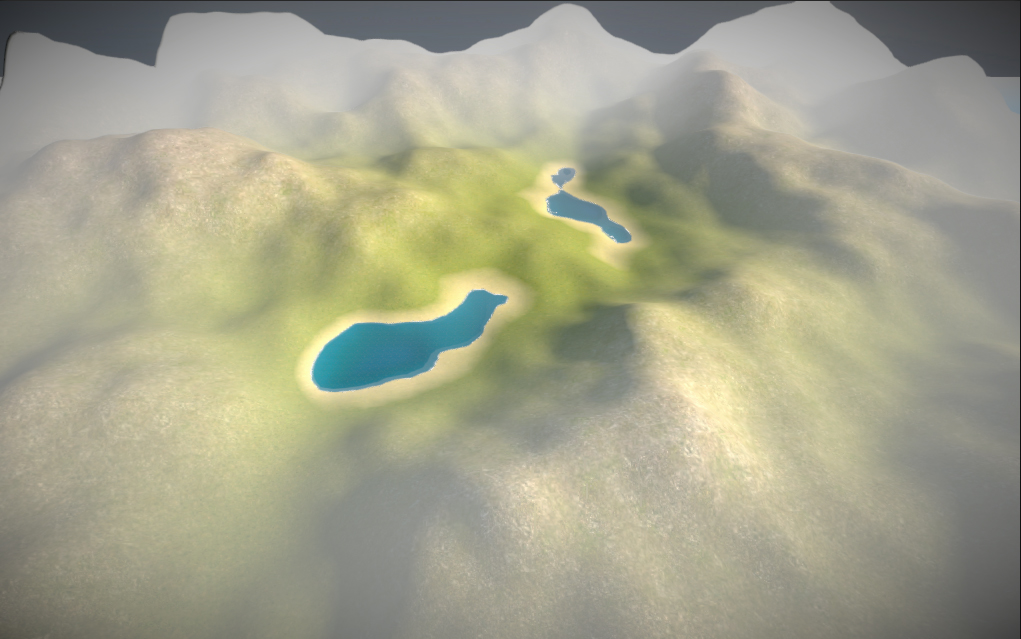}
\caption{Terrain}
\label{fig:env-terrain}
\end{subfigure}
\begin{subfigure}{0.19\textwidth}
\centering
\includegraphics[width=\textwidth]{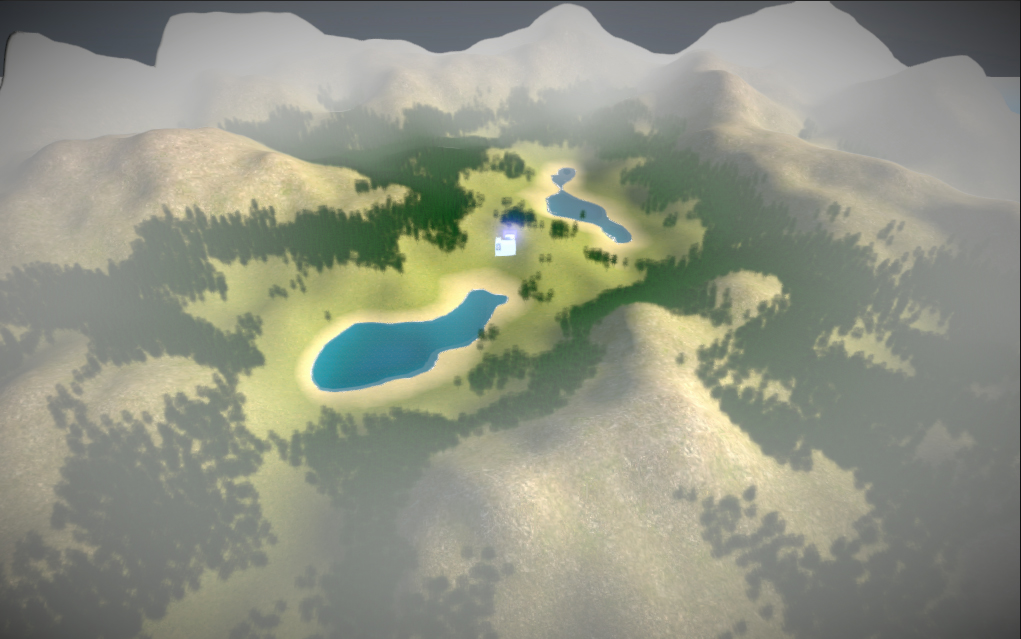}
\caption{Forest}
\label{fig:env-forest}
\end{subfigure}
\begin{subfigure}{0.19\textwidth}
\centering
\includegraphics[width=\textwidth]{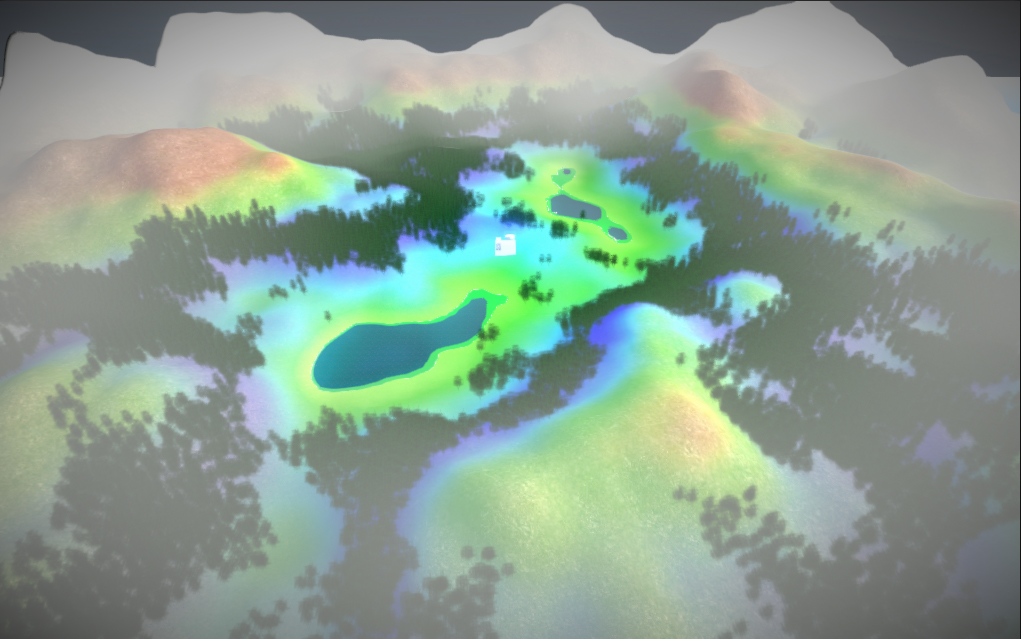}
\caption{O. Reforestation}
\label{fig:env-forest-noise}
\end{subfigure}
\begin{subfigure}{0.19\textwidth}
\centering
\includegraphics[width=\textwidth]{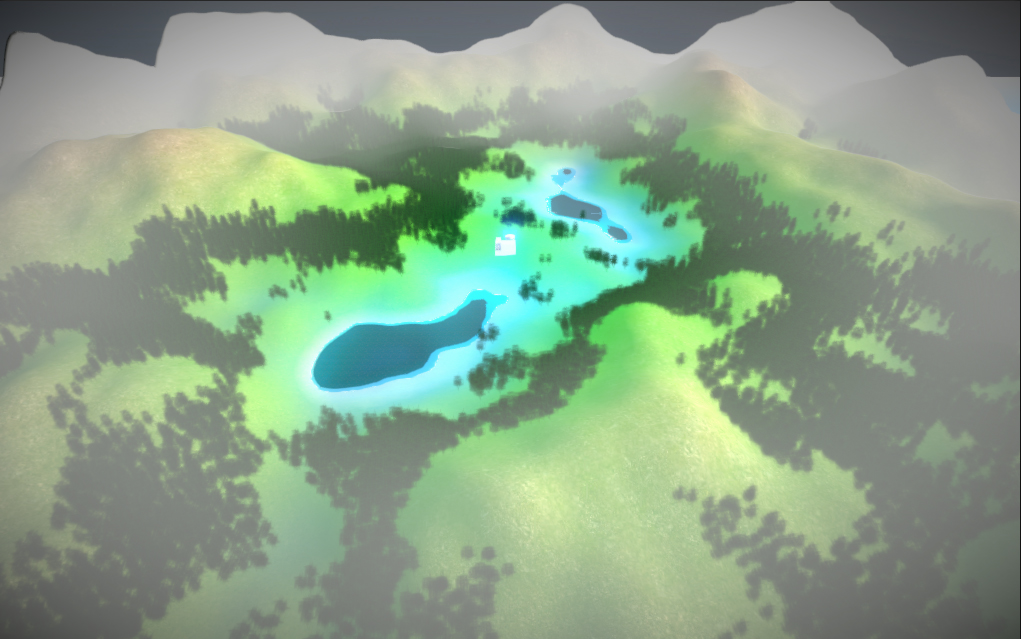}
\caption{Height Map}
\label{fig:env-height-map}
\end{subfigure}
\begin{subfigure}{0.19\textwidth}
\centering
\includegraphics[width=\textwidth]{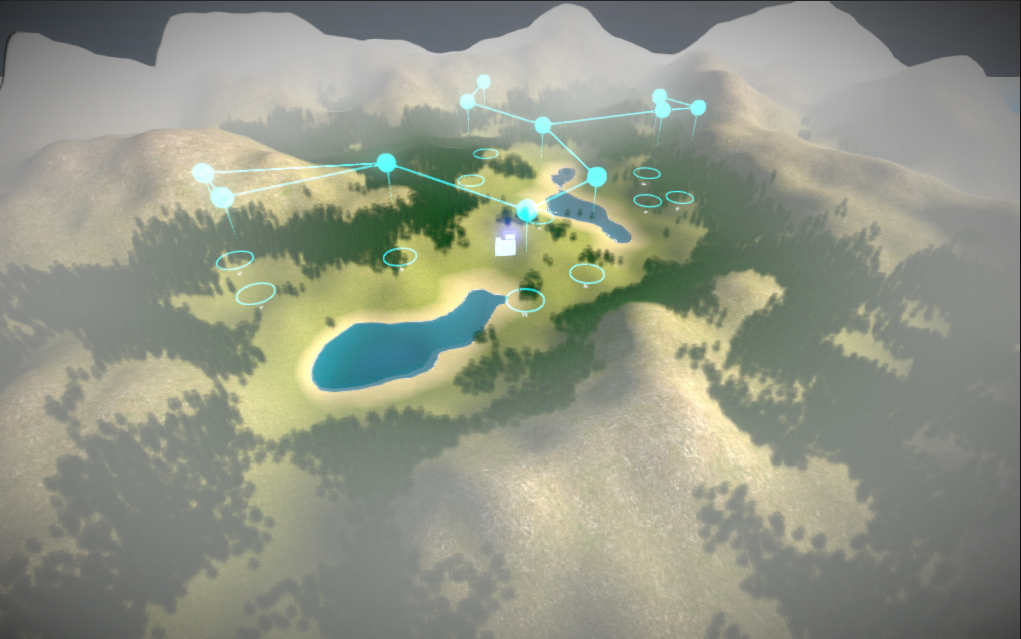}
\caption{Com. Network}
\label{fig:env-network}
\end{subfigure}
\caption{Static environment features. Zoomed-in version in the appendix: \ref{appendix:env-features-zoom}.}
\end{figure}

\begin{figure}[!ht]
\begin{subfigure}{0.19\textwidth}
\centering
\includegraphics[width=\textwidth]{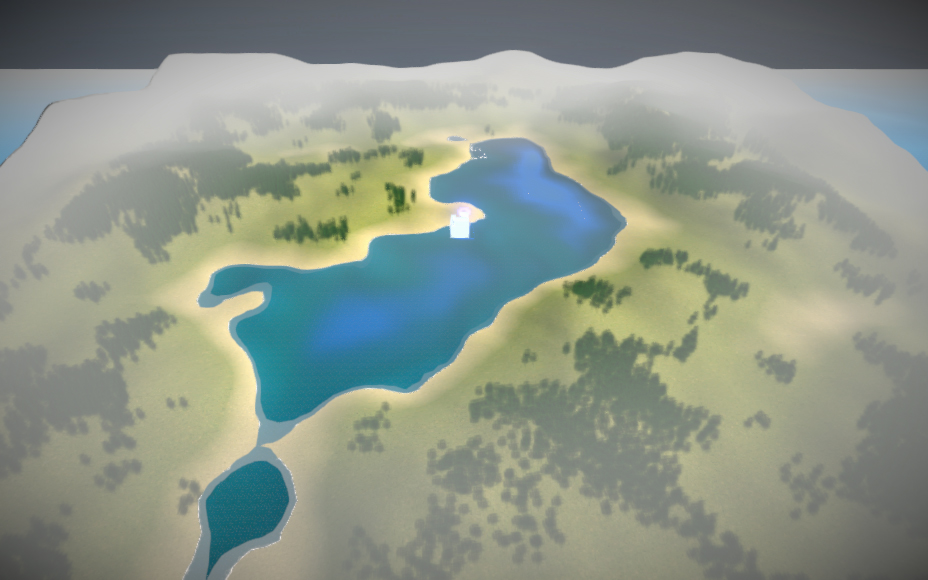}
\caption{Difficulty 1}
\label{fig:scenario-diff-0}
\end{subfigure}
\begin{subfigure}{0.19\textwidth}
\centering
\includegraphics[width=\textwidth]{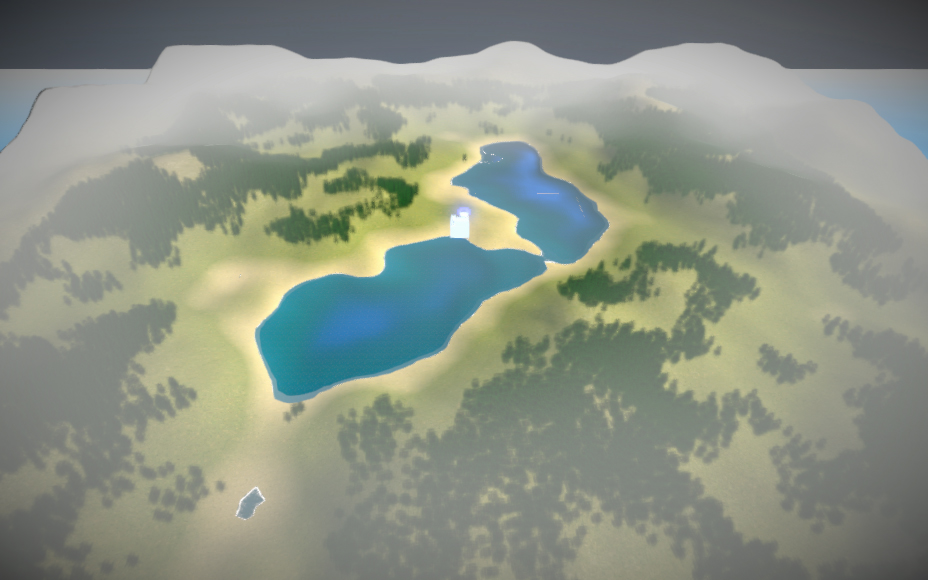}
\caption{Difficulty 2}
\label{fig:scenario-diff-1}
\end{subfigure}
\begin{subfigure}{0.19\textwidth}
\centering
\includegraphics[width=\textwidth]{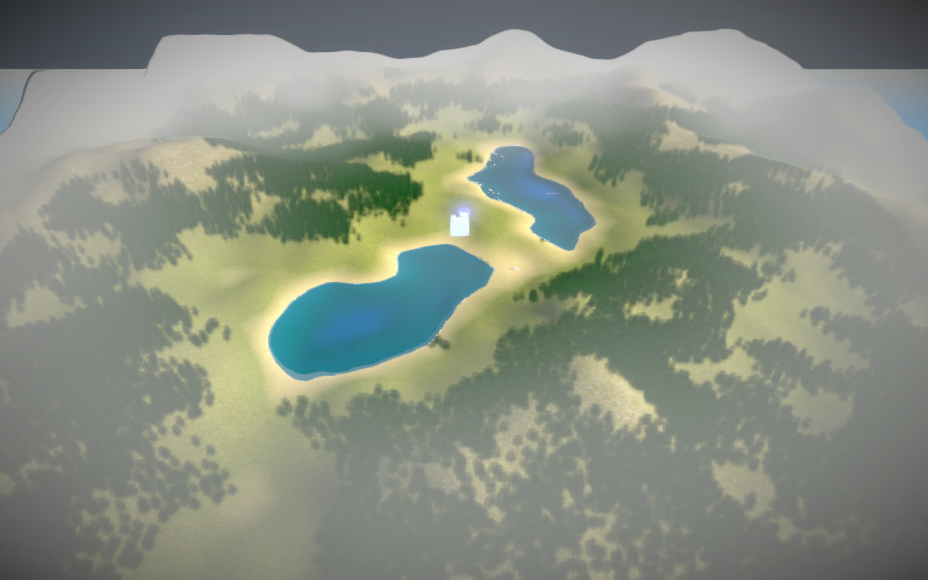}
\caption{Difficulty 3}
\label{fig:scenario-diff-2}
\end{subfigure}
\begin{subfigure}{0.19\textwidth}
\centering
\includegraphics[width=\textwidth]{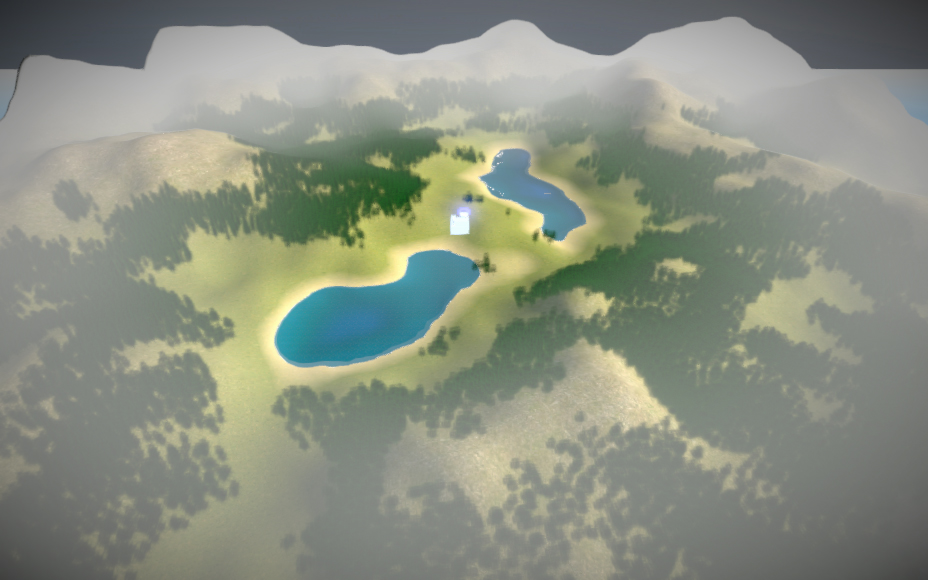}
\caption{Difficulty 4}
\label{fig:scenario-diff-3}
\end{subfigure}
\begin{subfigure}{0.19\textwidth}
\centering
\includegraphics[width=\textwidth]{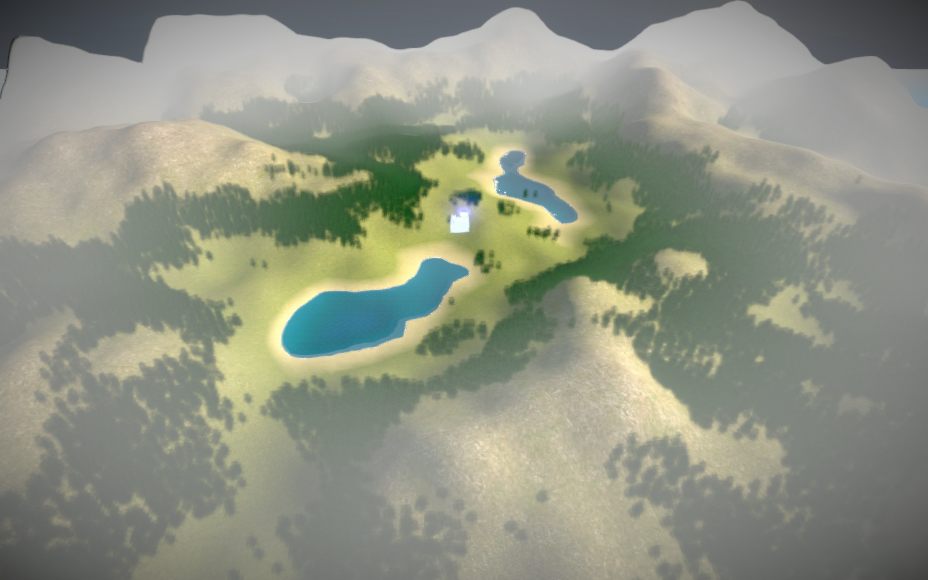}
\caption{Difficulty 5}
\label{fig:scenario-diff-4}
\end{subfigure}
\caption{Various difficulty terrain samples. Zoomed-in version in the appendix: \ref{appendix:env-difficulty-zoom}.}
\label{fig:env-difficulty}
\end{figure}

\begin{figure}[!ht]
\begin{subfigure}{0.19\textwidth}
\centering
\includegraphics[width=\textwidth]{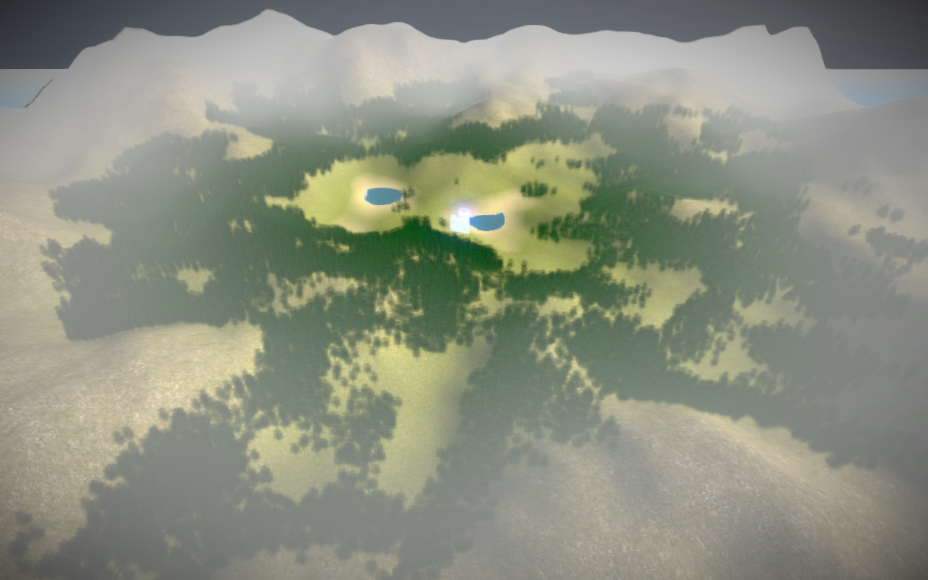}
\caption{Seed 25}
\label{fig:scenario-seed-25}
\end{subfigure}
\begin{subfigure}{0.19\textwidth}
\centering
\includegraphics[width=\textwidth]{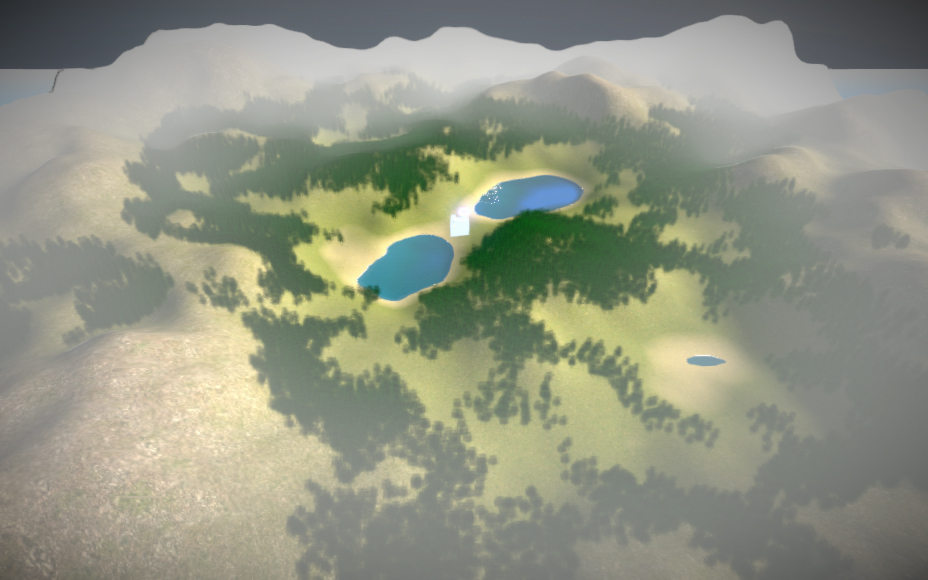}
\caption{Seed 26}
\label{fig:scenario-seed-26}
\end{subfigure}
\begin{subfigure}{0.19\textwidth}
\centering
\includegraphics[width=\textwidth]{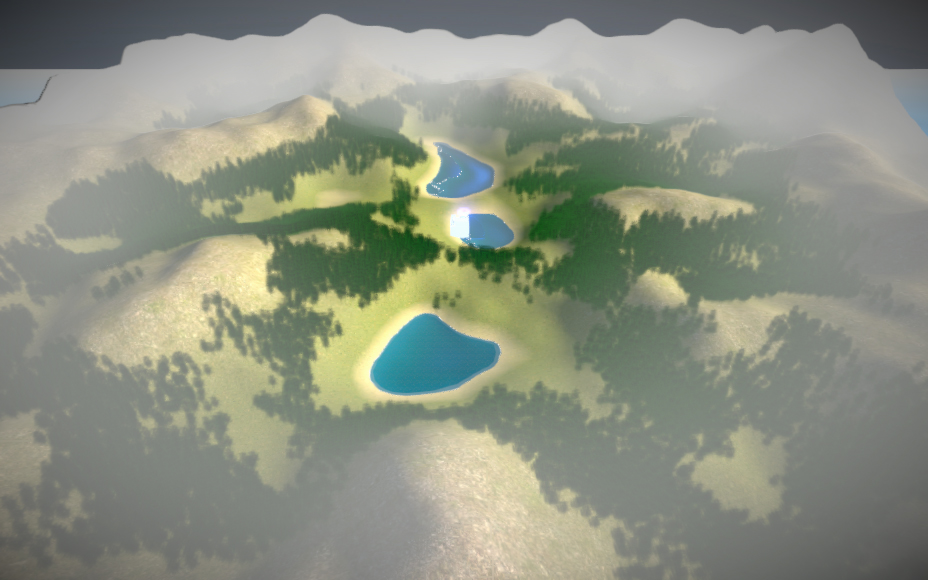}
\caption{Seed 27}
\label{fig:scenario-seed-27}
\end{subfigure}
\begin{subfigure}{0.19\textwidth}
\centering
\includegraphics[width=\textwidth]{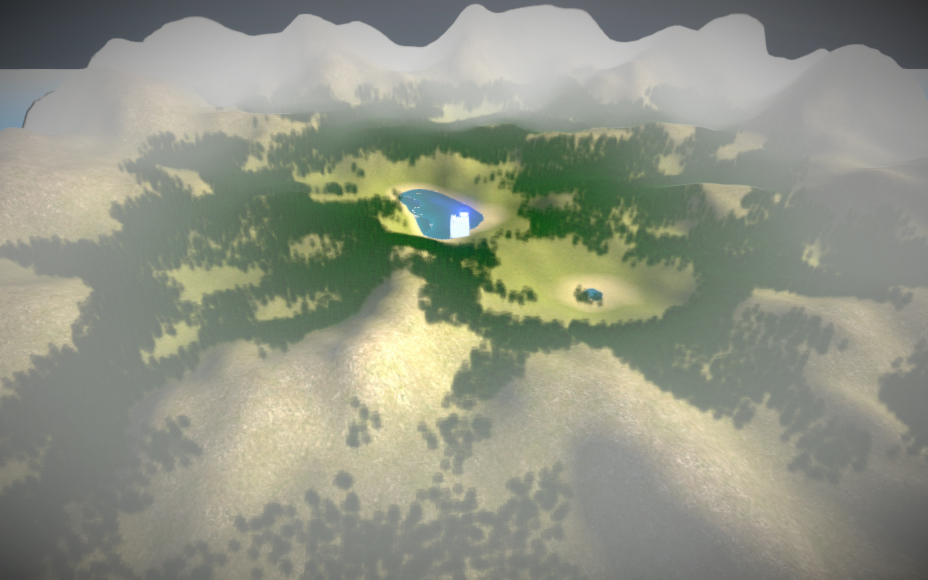}
\caption{Seed 28}
\label{fig:scenario-seed-28}
\end{subfigure}
\begin{subfigure}{0.19\textwidth}
\centering
\includegraphics[width=\textwidth]{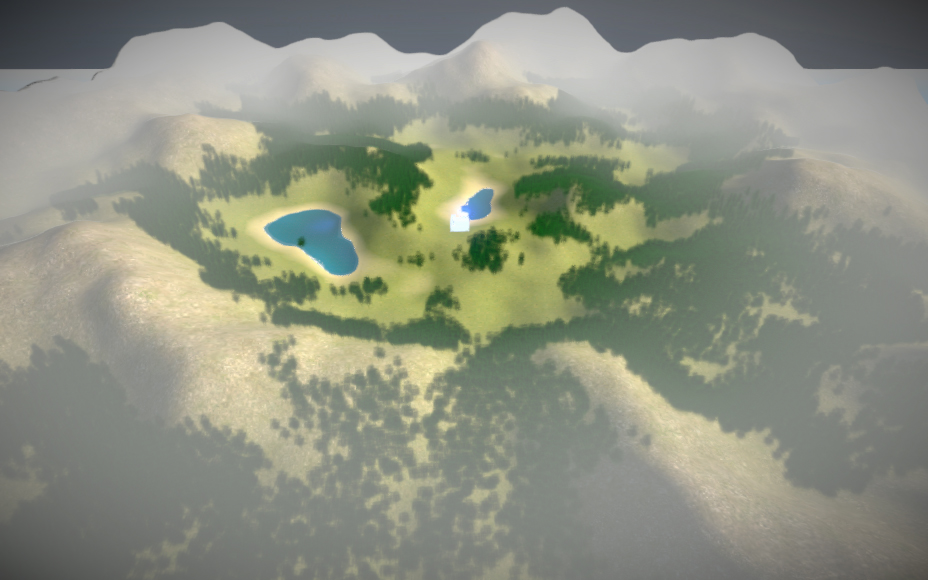}
\caption{Seed 31}
\label{fig:scenario-seed-31}
\end{subfigure}
\caption{Various random seed terrain scenarios. Zoomed-in version in the appendix: \ref{appendix:env-scenarios-zoom}.}
\label{fig:env-scenarios}
\end{figure}

\begin{figure}
\includegraphics[width=\linewidth]{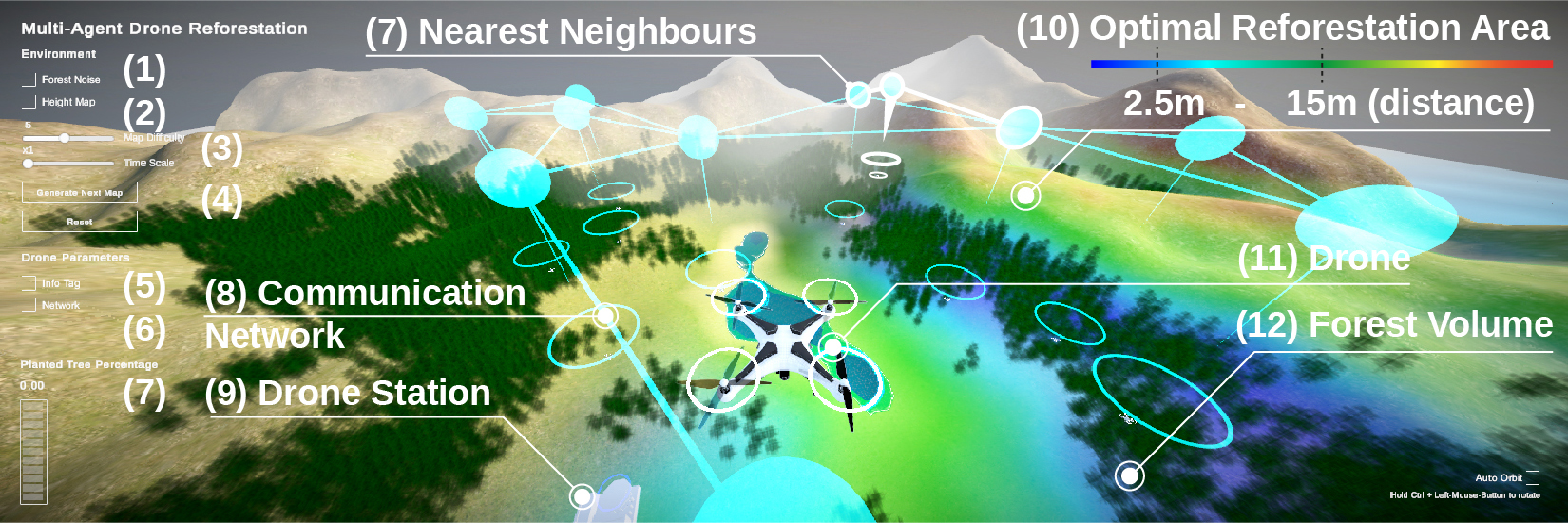}
\caption{Drone Reforestation Environment: (1) Forest Noise Toggle, (2) Height Map Toggle, (3) Terrain Difficulty Slider, (4) Time Scale Slider, (5) Info Tag Toggle, (6) Network Toggle, (7) Cumulative Performance Bar, (8) Communication Network, (9) Drone Station and Spawn Point, (10) Optimal Reforestation Area Heat Map, (11) Drone, (12) Forest Volume.}
\label{fig:environment}
\end{figure}

We now describe the 3-D Reforestation environment, developed in the game engine unity, used for simulation and online training of the Multi-Agent (MA) baseline and our Multi-Agent communication setup (MAC). The environment is considered solved if an agent picks up a tree seed from the drone station, finds a spot ideal for planting a tree seed, drops the tree seed and returns to the drone station to recharge the battery and pick up the next tree seed. All variations of environment scenarios consist of the following elements: 1. Procedural generated terrain, using octaves, persistence and lacunarity-based noise \ref{fig:env-terrain}, 2. The forest trees, placed in a certain height region, combined with a random noise map, on fertile ground only, here displayed as green grass \ref{fig:env-forest}, 3. The human user interface allows to display the regions that are best for reforestation and yield the highest amount of reward; however, this is not visible to the agent's \ref{fig:env-forest-noise}, 4. A height heat map, also only visible in the human user interface \ref{fig:env-height-map} and 5. A network that defines the nearest neighbours dynamically displayed in cyan, visible in the human user interface only \ref{fig:env-network}. Part of the environment, acting as spawn, tree seed- and battery-charging point, is a drone station marked with (9) on Figure \ref{fig:environment}. Ten drones controlled by individual agents are spawning at the drone station spawn point in all experiments. The terrain in all scenarios spans 1200 meters in both directions and has a maximum mountain altitude of 100 meters, depending on the difficulty level. Various terrain samples of difficulty levels 1-5 are shown in Figure \ref{fig:env-difficulty}. Furthermore, an additional bowl-shaped height filter adjusts the terrain to be close to a valley when difficulty levels increase. We can generate open-ended scenarios with a random seed input - sample terrains of various random seeds are shown in Figure \ref{fig:env-scenarios}.

\subsection{Agent Setup}

\textbf{Goal:} Each agent must learn to navigate the drone to the drone station spawn point, where it is automatically serviced, picks up a tree seed and recharges the battery. It then has to fly the drone and scout an ideal spot for dropping the tree seed held while keeping track of battery life, which also needs to cover the way back to the drone station.\\
\textbf{Reward Function:} The agent reward function consists of multiple parts. If the drone does not hold a tree seed, the agent must navigate back to the drone station. Incremental rewards are yielded for getting closer to the drone station, accumulating to a total of +20, independent of the distance. The agent receives the last increment for arriving at the drone station, picking up a tree seed and recharging the battery. Finally, dropping a tree seed yields a reward of +0 to +30, depending on the location. The ideal tree seed drop area is defined by the distance to existing trees. The reward of +0 to +30 is mapped to a distance as close as 75 to 2.5 meters. I.e., if a tree seed is dropped as close as 2.5 meters from an existing tree, the reward yielded is +30, for a tree seed drop distance of 10 meters, the reward is +26.8, and a tree seed dropped at a distance below 2.5 or above 75 meters yields a reward of +0. The total reward of +30 includes a distance bonus of +10. If the agent is risking running out of battery and finds an excellent spot for a tree seed drop, the distance between the tree seed drop location and the drone station spawn point is factored into the total reward. The battery life ranges from a full charge of 1 and empty at 0. For each step, while holding a tree seed - a higher payload - the battery is being depleted at a rate of -0.001, holding no tree seed at a rate of -0.0005, and yields negative rewards accordingly. This results in 1000 environment steps with a tree seed and 2000 steps without a tree seed until running out of battery life. The gold standard receives a maximum reward of +50 for a single task, excluding negative payload rewards.\\
\textbf{Observations:} The observation space consist of vector and visual observations.\\ \textbf{Vector Observations:} All vector observations are normalized and consists of the following: The distance from the drone to the ground as float [0-100], the location of the drone in 3-D space as a vector(x, y, z) [-600-600], the movement direction vector(x, y, z) [0-1], the vector from the drone to the drone station spawn point as a vector(x, y, z) [-600-600], if the drone is holding a tree seed as a bool [true / false] mapped to [0, 1], the battery status as a float [0-1] and lastly three inbox spaces to receive messages from the neighbouring drone's memory, which consist of vector(x, y, z) [-600-600] location information. The final vector observation space size is 30, consisting of 2 stacks of the 15 described observations.\\
\textbf{Visual Observations:} The visual observation is a grey scale grid of 16x16, cells captured by a down-ward pointing camera attached to each drone with a field of view of 120 - 256 cells each with a float value ranging from [0-1]. This results in a total observation space size of 286. We use a residual neural network (ResNet) architecture consisting of three stacked layers, each with two residual blocks \citep{espeholt_impala_2018} for our visual observations processing.\\
\textbf{Actions:} The action space consists of a combination of continuous and discrete actions.\\
\textbf{Continuous Actions:} Each agent has three continuous actions with values in the range of -1 to 1. Continuous actions control the movement of the drone. Continuous action 0 controls the forward and backward movement, action 1 controls the rotation, left and right, and action 2 controls the up and down movement of the drone. The movement speed of the drone is 1-meter per time-step, and the rotation speed is 5-degree per time-step.\\
\textbf{Discrete Actions:} The discrete action space size is two and can be described as a tree with two branches, each with two possible values [0, 1]. Discrete action 0 drops the tree seed at value 1, and action 1 saves a location to memory at value 1. The location memory can hold a three-dimensional vector representing a location in the environment. Both actions do nothing if the value is at 0.

\subsection{Multi-Agent Communication}

\begin{figure}
\includegraphics[width=\linewidth]{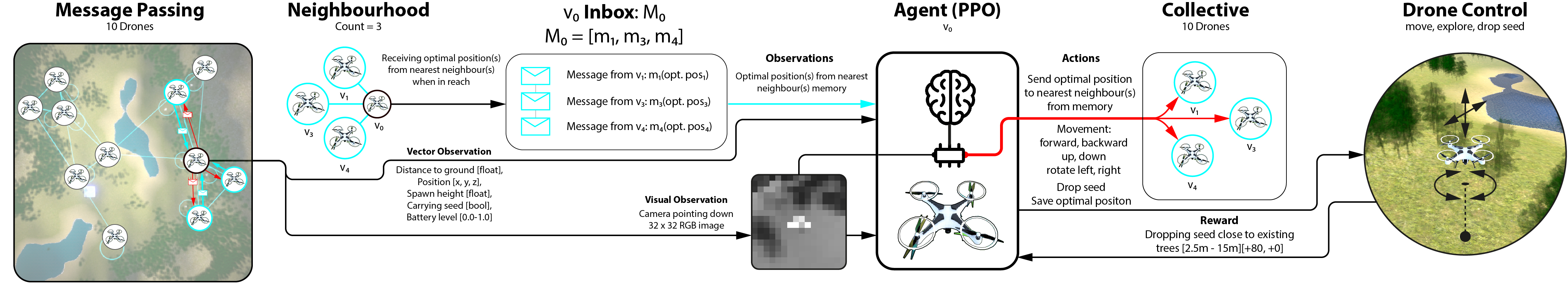}
\caption{GNN Message Passing Diagram: Message Passing across neighbourhood; Inbox of the individual Drone; Locations for optimal tree seed drop, as part of the RL Agent observation space.}
\label{fig:GNN_MessagePassing}
\end{figure}

Our learning mechanism allows the agent to receive graph-structured data. Messages can be exchanged if a drone is within 200 meters of another drone. A total number of three messages can be received, corresponding to three nearest neighbours in the Multi-Agent communication setup (MAC). If there are only two drones close enough, only two messages are received and sent respectively. Sending or receiving a message has no cost and yields no negative or positive reward.

\section{Experiments}

\begin{table}[H]
\vspace{-0.5cm}
\caption{Experiment Setup}
\label{experiment-table}
\centering
\begin{tabularx}{\textwidth}{p{5.4cm} p{1.3cm} p{1.8cm} p{2.0cm} p{1.4 cm}}
\toprule
Experiment&Agent(s)&Neighbour(s)&Training Seed&Test Seed\\
\midrule
1 Multi-Agent (MA)           & 10            & 0                 & 0                 & 111\\
2 Multi-Agent (MA)    & 10            & 0                 & 0-99              & 111\\
\midrule
3 Multi-Agent Communication (MAC)         & 10            & 3                 & 0                 & 111\\
4 Multi-Agent Communication (MAC) & 10            & 3                 & 0-99              & 111\\
\bottomrule
\end{tabularx}
\end{table}

\begin{figure}[H]
    \begin{subfigure}{0.245\textwidth}
    \centering
    \includegraphics[width=\textwidth]{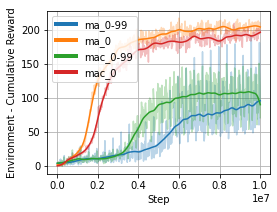}
    \caption{Cumulative Reward}
    \label{fig:train-results-cum-reward}
    \end{subfigure}
    \begin{subfigure}{0.245\textwidth}
    \centering
    \includegraphics[width=\textwidth]{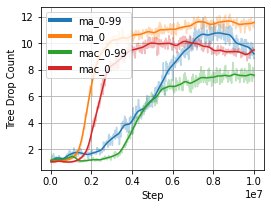}
    \caption{Tree Drop Count}
    \label{fig:train-results-tree-drop-count}
    \end{subfigure}
    \begin{subfigure}{0.245\textwidth}
    \centering
    \includegraphics[width=\textwidth]{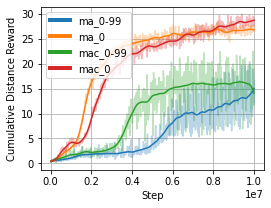}
    \caption{Cml. Distance Reward}
    \label{fig:train-results-cum-dist-reward}
    \end{subfigure}
    \begin{subfigure}{0.245\textwidth}
    \centering
    \includegraphics[width=\textwidth]{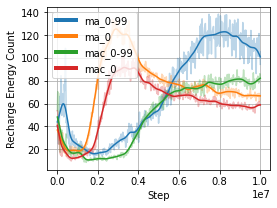}
    \caption{Recharge Energy Count}
    \label{fig:train-results-recharge-energy}
    \end{subfigure}
\caption{Training Graphs. Zoomed-in version and additional data in the appendix: \ref{appendix:train-results-graphs-zoomed}.}
\label{fig:train-results-graphs}
\end{figure}

We have trained four different setups for our experiments (Figure \ref{fig:train-results-graphs}). \textbf{Multi-Agent Setup without Communication as Baseline:} Experiment 1 and 2 have been trained without the ability to communicate: Experiment 1 is trained on the terrain scenario with the random seed 0 and Experiment 2 on the terrain scenarios with random seeds ranging from 0 to 99 in a sequence. \textbf{Multi-Agent Setup with Communication:} Experiment 3 and 4 have been trained with the ability to communicate: Experiment 3 is trained on the terrain scenario with the random seed 0 and Experiment 4 on the terrain scenarios with random seeds ranging from 0 to 99 in a sequence. All experiments are then tested on an unseen terrain scenario with the random seed 111.

\section{Results and Discussion}

\begin{figure}[H]
    \vspace{-0.5cm}
    \begin{subfigure}{0.24\textwidth}
    \centering
    \includegraphics[width=\textwidth]{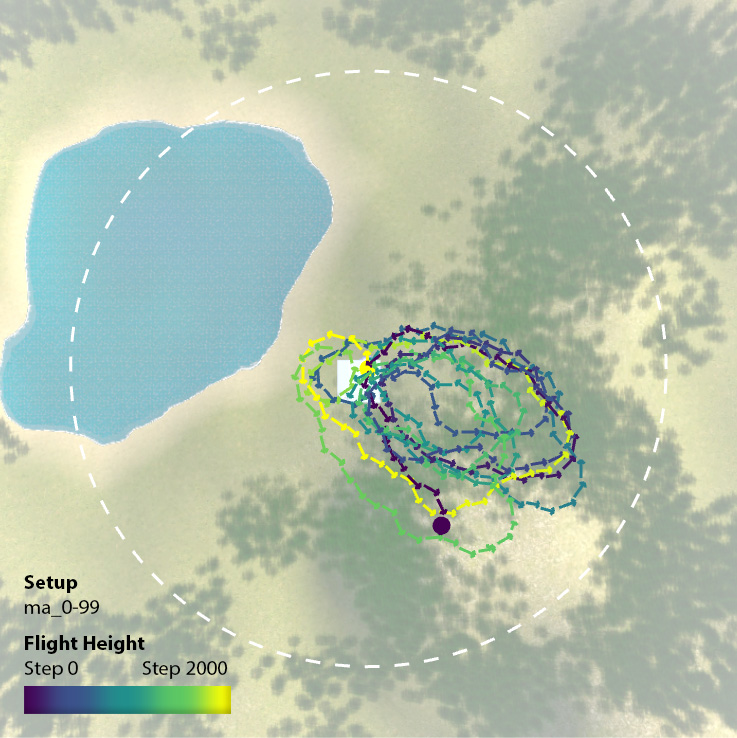}
    \caption{MA 0-99 Flight Path}
    \label{fig:ma-0-99-flight-path}
    \end{subfigure}
    \begin{subfigure}{0.24\textwidth}
    \centering
    \includegraphics[width=\textwidth]{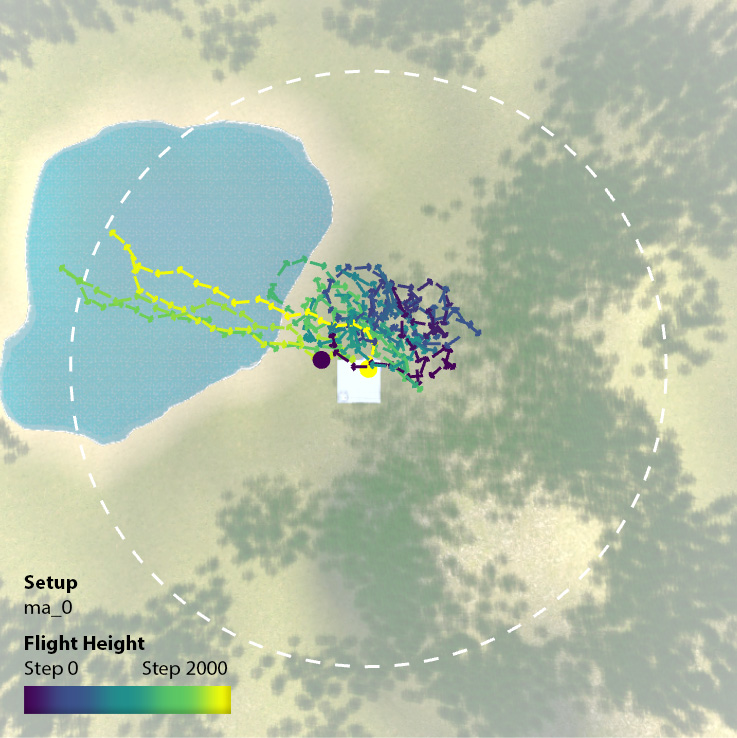}
    \caption{MA 0 Flight Path}
    \label{fig:ma-0-flight-path}
    \end{subfigure}
    \begin{subfigure}{0.24\textwidth}
    \centering
    \includegraphics[width=\textwidth]{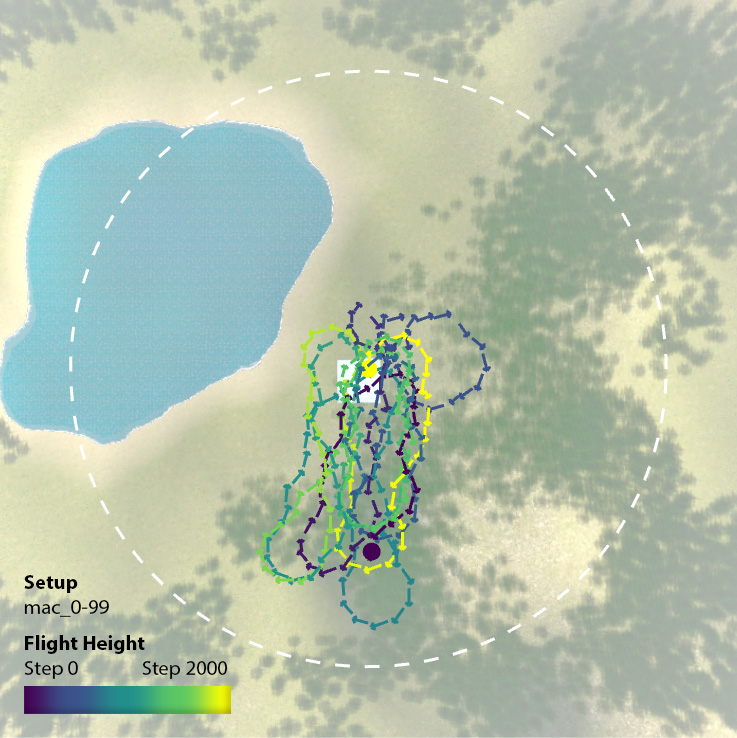}
    \caption{MAC 0-99 Flight Path}
    \label{fig:mac-0-99-flight-path}
    \end{subfigure}
    \begin{subfigure}{0.24\textwidth}
    \centering
    \includegraphics[width=\textwidth]{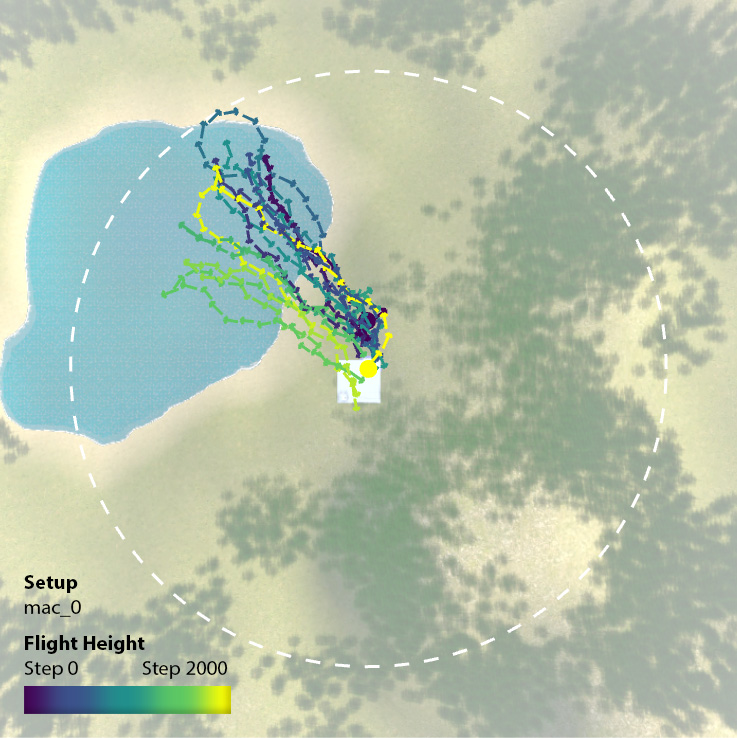}
    \caption{MAC 0 Flight Path}
    \label{fig:mac-0-flight-path}
    \end{subfigure}
\caption{2000 Step sample flight path (terrain seed 111). Zoomed-in version in the appendix: \ref{appendix:inference-results-graphs-zoomed}.}
\label{fig:results-flight-paths}
\end{figure}

Our results (Figure \ref{results-table}) show that Multi-Agent communication outperforms our non-communication Multi-Agent baseline setup. The MA 0 setup that has been trained on the terrain scenario with a random seed 0 performs very poorly on the test terrain scenario with the random seed 111. While the tree drop count is high, the agent drops tree seeds very conservatively without exploring and therefore receives a low cumulative reward. The Multi-Agent setup with the ability to communicate that has also been trained on the terrain scenario with the random seed 0, MAC 0, performs similarly bad, but through communication starts to explore marginally more and therefore receives a higher distance reward. Nevertheless, the cumulative reward is low as well. In contrast, the MA 0-99 setup, that has been trained on terrain scenario with a random seed ranging from 0-99, performs marginally the best in regards to the cumulative reward. Agents trained on multiple scenarios perform better, by a large margin, in comparison to agents that have been trained on a single terrain scenario. While the MA 0-99 setup has the highest cumulative reward marginally, we can observe that the MAC 0-99 setup, achieves the highest tree drop count and travels the furthest to explore (Figure \ref{fig:results-flight-paths}), through communication.

\begin{table}[H]
\caption{Experiment results: Mean Cumulative Reward, Mean Cumulative Tree Drop Count, Cumulative Reward. Testing terrain scenario random seed: 111. Each tested 10 times for 1e6 Steps.}
\label{results-table}
\centering
\begin{tabularx}{\textwidth}{p{1.8cm} p{3.7cm} p{3.7cm} p{3.7cm}}
\toprule
Experiment & Distance Reward (↑ better) & Tree Drop Count (↑ better) & Cml. Reward (↑ better) \\
\midrule
1 MA 0           & 1.29 (±0.40)      & 7.75 (±0.53)              & 2.54 (±1.23) \\
2 MA 0-99      & 11.45 (±0.95)       & 7.91 (±0.46)               & \textbf{122.08 (±8.7)} \\
\midrule
3 MAC 0          & 2.85 (±0.78)       & 6.61 (±0.51)              & 7.82 (±2.66) \\
4 MAC 0-99    & \textbf{13.18 (±1.14)}  & \textbf{8.66 (±0.46)}      & 121.84 (±8.79) \\
\bottomrule
\end{tabularx}
\end{table}

\begin{figure}[H]
    \vspace{-0.5cm}
    \begin{subfigure}{0.245\textwidth}
    \centering
    \includegraphics[width=\textwidth]{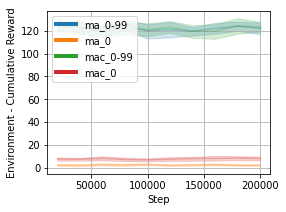}
    \caption{Cumulative Reward}
    \label{fig:test-results-cum-reward}
    \end{subfigure}
    \begin{subfigure}{0.245\textwidth}
    \centering
    \includegraphics[width=\textwidth]{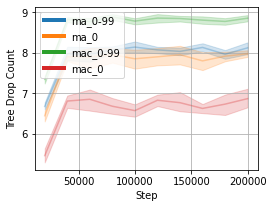}
    \caption{Tree Drop Count}
    \label{fig:test-results-tree-drop-count}
    \end{subfigure}
    \begin{subfigure}{0.245\textwidth}
    \centering
    \includegraphics[width=\textwidth]{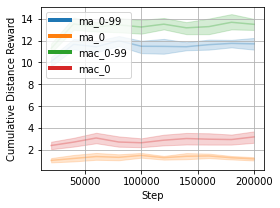}
    \caption{Cml. Distance Reward}
    \label{fig:test-results-cum-dist-reward}
    \end{subfigure}
    \begin{subfigure}{0.245\textwidth}
    \centering
    \includegraphics[width=\textwidth]{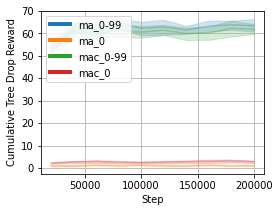}
    \caption{Cml. Tree Drop Reward}
    \label{fig:test-results-cum-tree-drop-reward}
    \end{subfigure}
\caption{Test Graphs. Zoomed-in version and additional data in the appendix: \ref{appendix:inference-results-graphs-zoomed}.}
\label{fig:test-results-graphs}
\end{figure}

\section{Conclusion}

This work is an approach to solving autonomous drone-based reforestation using Multi-Agent Reinforcement Learning (MARL) with a Graph Neural Network (GNN) communication layer that enables agents to collaborate. We have demonstrated that we can solve this task and generalise well on an unseen terrain scenario. We also show that communication can lead to collaboration and increase the performance of a multi-agent collective, ultimately outperforming the multi-agent setup without the ability to communicate. This is verified by increased tree seed drop counts and the quality and precision of tree seed planting, yielding higher rewards and, subsequently, a more efficient reforestation. Furthermore, we discovered that communication can lead to a higher risk-taking propensity and a larger area of forest explored. If an agent made it to a specific forest area, other agents can. We understand that the simulation to reality gap still exists, but this is a first step toward approaching this high-impact problem. We can see multiple ways to push this work forward. Firstly, improving the environment to be more realistic, with photo-realistic textures, trees, landscape, vegetation and real-world terrain data. The implemented drone control is an abstraction of a drone controller. Secondly, more realistic physics, including winds, and a control system that is closely aligned with standard drone controls, i.e. pitch, yaw and roll. And lastly, the camera vision and sensing can be improved. In this work, we use a 16x16 image as a visual observation input; this could be increased, and subsequently, the network architecture for processing. Furthermore, instead of a grey-scale image, we can add a RGB channel and process the observation image to additionally forward i.e. a depth map. This could lead to higher action precision and improved generalisability. Nevertheless, we believe we contributed value to the multi-agent reinforcement learning, collaboration and communication field on the back of a high-impact problem of relevance and urgency.


\newpage

\begin{ack}
We want to thank Jasmin Arensmeier, without her constant support, patience, guidance and
encouragement this would not have been possible. We also want to thank the reviewing committee for their efforts and critic. Further information, video material and an interactive web-app can be found at: \url{https://ai.philippsiedler.com/neurips2022-cooperativeai-gnn-marl-autonomous-drone-reforestation/}.
\end{ack}

{
\bibliography{neurips_2022}
}

\appendix

\section{Appendix}

\section{Pseudocode}

PPO-CLIP pseudocode \citep{openai_proximal_2021, schulman_proximal_2017}:

\begin{algorithm}
	\caption{PPO-Clip}
	
	\begin{algorithmic}[1]
		\item Input: initial policy parameters $\theta_0$, initial value function parameters $\phi_0$
		\For {$k=0,1,2,\ldots$}
		    \State Collect set of trajectories $\mathcal{D}_k$ = \{$\tau_i$\} by running policy $\pi_k = \pi(\theta_k)$ in the environment.
		    \State Compute rewards-to-go $\hat{R_t}$.
		    \State Compute advantage estimates, $\hat{A_t}$ (using any method of     advantage estimation) based on the
		    \State current value function $V_{\phi_k}$
			\State Update the policy by maximizing the PPO-Clip objective:
			\State $\theta_{k+1} = arg\underset{\theta}{max} \frac{1}{|\mathcal{D}_k|T} \sum_{\tau \in \mathcal{D}_k} \sum_{t = 0}^{T} \min \left( \frac{\pi_\theta(a_t|s_t)}{\pi_{\theta_k}(a_t|s_t)}A^{\pi_{\theta_k}}(s_t, a_t), g(\epsilon, A^{\pi_{\theta_k}}(s_t, a_t)) \right)$,
			\State typically via stochastic gradient ascent with Adam.
			\State Fit value function by regression on mean-squared error:
			\State $\phi_{k+1} = arg\underset{\phi}{min} \frac{1}{|\mathcal{D}_k|T} \sum_{\tau \in \mathcal{D}_k} \sum_{t = 0}^{T} \left( (V_{\phi}(s_t)-\hat{R_t} \right)$
			\State typically via some gradient descent algorithm.
		\EndFor
	\end{algorithmic} 
\end{algorithm}

Simple Multi-Agent PPO pseudocode:

\begin{algorithm}
	\caption{Multi-Agent PPO} 
	\begin{algorithmic}[1]
		\For {$iteration=1,2,\ldots$}
			\For {$actor=1,2,\ldots,N$}
				\State Run policy $\pi_{\theta_{old}}$ in environment for $T$ time steps
				\State Compute advantage estimates $\hat{A}_{1},\ldots,\hat{A}_{T}$
			\EndFor
			\State Optimize surrogate $L$ wrt. $\theta$, with $K$ epochs and minibatch size $M\leq NT$
			\State $\theta_{old}\leftarrow\theta$
		\EndFor
	\end{algorithmic} 
\end{algorithm}

\newpage
\section{Hyperparameters}
\subsection{Multi Agent Training Hyperparameters}
\begin{verbatim}
behaviors:
  MA_Drone:
    trainer_type: ppo
    hyperparameters:
      batch_size: 1024
      buffer_size: 10240
      learning_rate: 0.0003
      beta: 0.005
      epsilon: 0.2
      lambd: 0.95
      num_epoch: 3
      learning_rate_schedule: linear
    network_settings:
      normalize: false
      hidden_units: 128
      num_layers: 2
      vis_encode_type: resnet
    reward_signals:
      extrinsic:
        gamma: 0.99
        strength: 0.9
        network_settings:
          vis_encode_type: resnet
      curiosity:
        gamma: 0.99
        strength: 0.1
        encoding_size: 256
        learning_rate: 0.0003
        network_settings:
          vis_encode_type: resnet
    keep_checkpoints: 5
    max_steps: 10000000
    time_horizon: 100
    summary_freq: 20000
    threaded: true
\end{verbatim}

\newpage
\subsection{Hyperparameter Description}

\begin{table}[h]
  \begin{tabular}{p{0.3\textwidth}p{0.3\textwidth}p{0.3\textwidth}}
    \toprule
    Hyperparameter & Typical Range & Description\\
    \midrule
    Gamma & $0.8-0.995$ & discount factor for future rewards\\
    Lambda & $0.9-0.95$ & used when calculating the Generalized Advantage Estimate (GAE)\\
    Buffer Size & $2048-409600$ & how many experiences should be collected before updating the model\\
    Batch Size & $512-5120$ (continuous), $32-512$ (discrete) & number of experiences used for one iteration of a gradient descent update.\\\
    Number of Epochs & $3-10$ & number of passes through the experience buffer during gradient descent\\
    Learning Rate & $1e-5-1e-3$ & strength of each gradient descent update step\\
    Time Horizon & $32-2048$ & number of steps of experience to collect per-agent before adding it to the experience buffer\\
    Max Steps & $5e5-1e7$ & number of steps of the simulation (multiplied by frame-skip) during the training process\\
    Beta & $1e-4-1e-2$ & strength of the entropy regularization, which makes the policy "more random"\\
    Epsilon & $0.1-0.3$ & acceptable threshold of divergence between the old and new policies during gradient descent updating\\
    Normalize & $true/false$ & weather normalization is applied to the vector observation inputs\\
    Number of Layers & $1-3$ & number of hidden layers present after the observation input\\
    Hidden Units & $32-512$ & number of units in each fully connected layer of the neural network\\
    \midrule
    Intrinsic Curiosity Module\\
    \midrule
    Curiosity Encoding Size & $64-256$ & size of hidden layer used to encode the observations within the intrinsic curiosity module\\
    Curiosity Strength & $0.1-0.001$ & magnitude of the intrinsic reward generated by the intrinsic curiosity module\\
    \bottomrule
  \end{tabular}
\end{table}

\newpage
\section{Environment Features}
\label{appendix:env-features-zoom}
\begin{figure}[H]
\begin{subfigure}{0.495\textwidth}
\centering
\includegraphics[width=\textwidth]{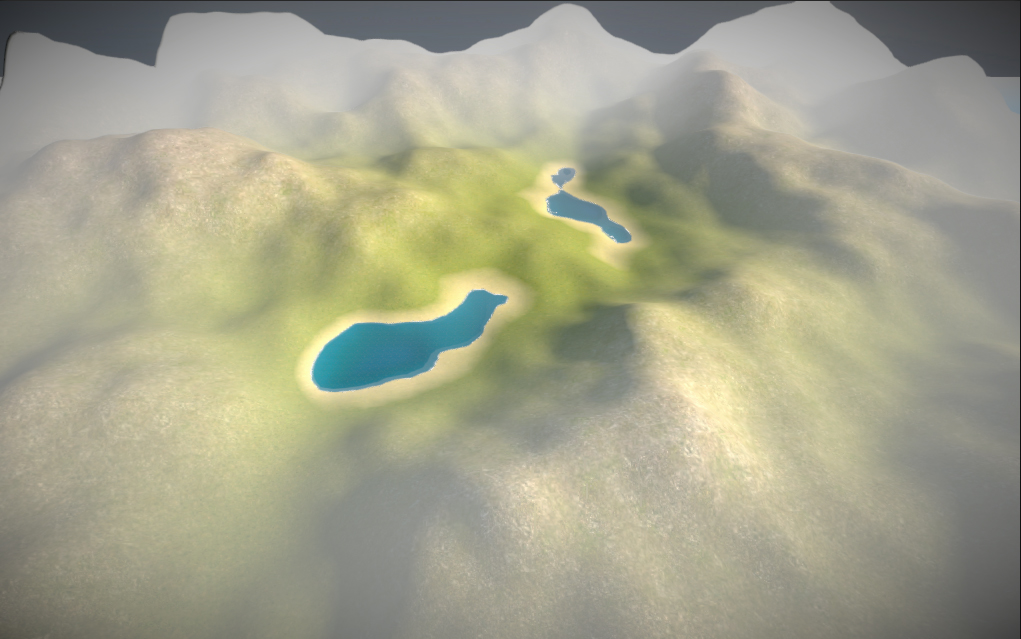}
\caption{Terrain, Seed 23, Difficulty 5}
\end{subfigure}
\begin{subfigure}{0.495\textwidth}
\centering
\includegraphics[width=\textwidth]{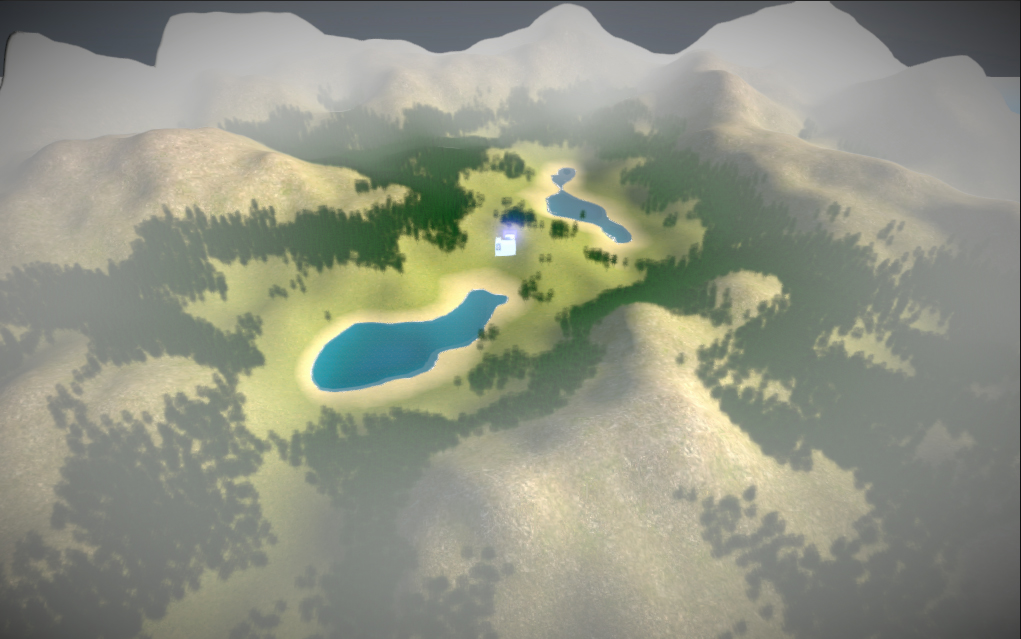}
\caption{Forest, Seed 23, Difficulty 5}
\end{subfigure}
\begin{subfigure}{0.495\textwidth}
\centering
\includegraphics[width=\textwidth]{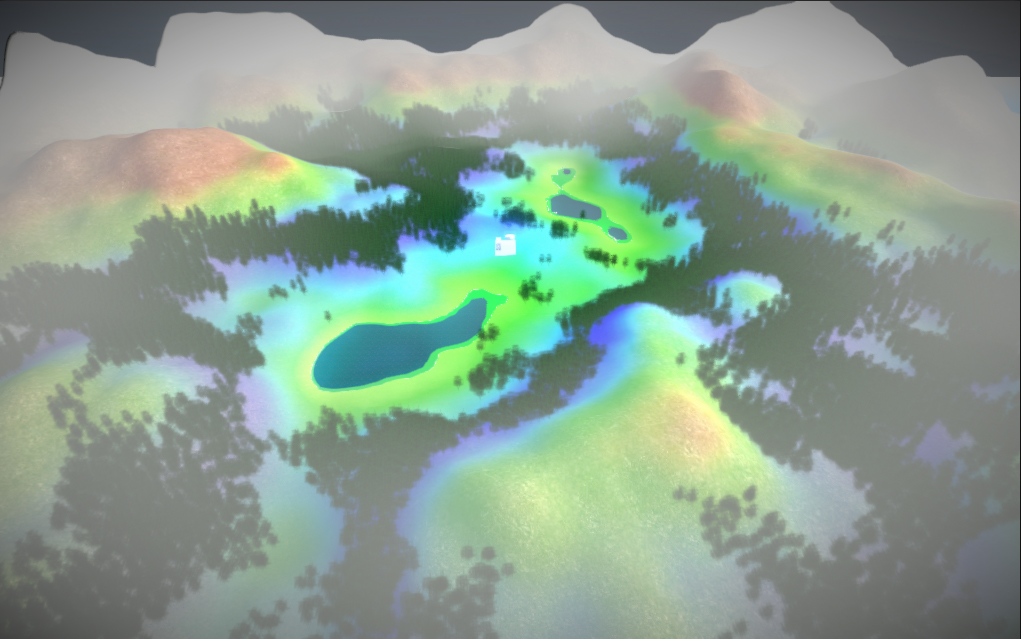}
\caption{Optimal Reforestation Map, Seed 23, Difficulty 5}
\end{subfigure}
\begin{subfigure}{0.495\textwidth}
\centering
\includegraphics[width=\textwidth]{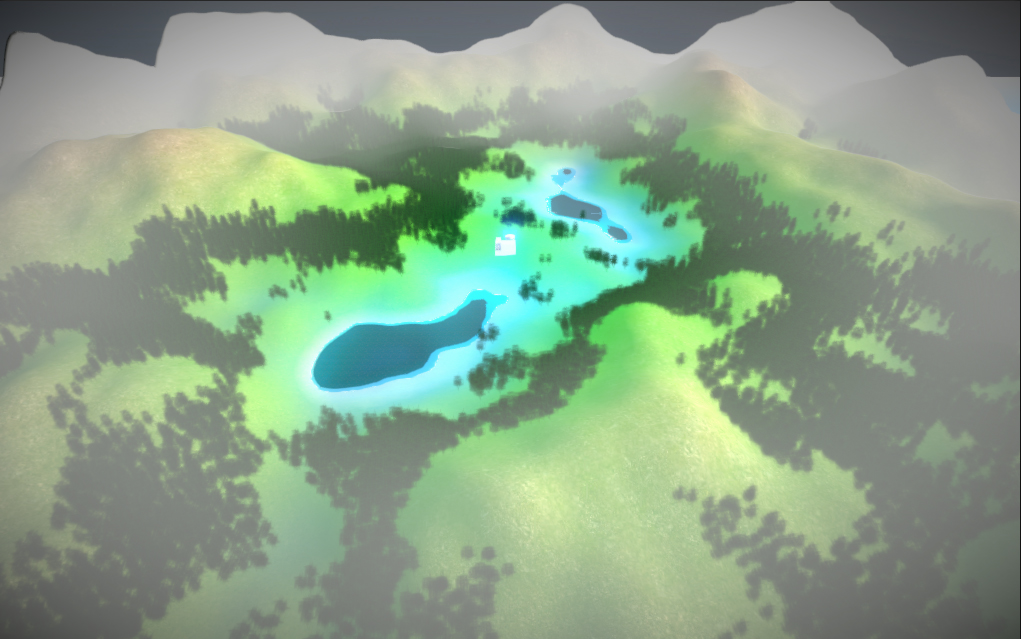}
\caption{Terrain Height Map, Seed 23, Difficulty 5}
\end{subfigure}
\begin{subfigure}{0.495\textwidth}
\centering
\includegraphics[width=\textwidth]{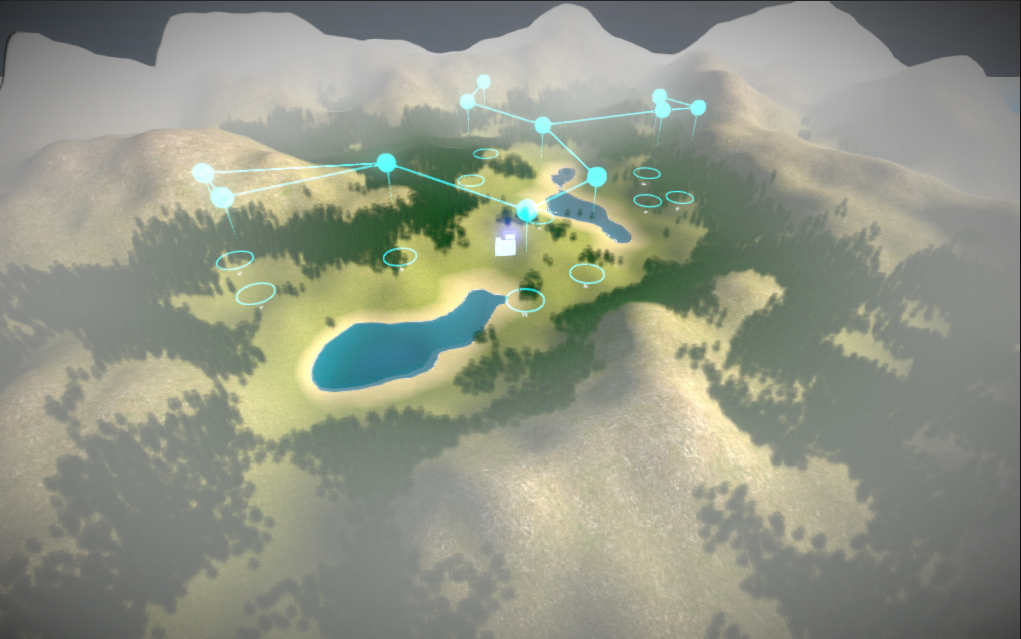}
\caption{Dynamic Drone Communication Network, Seed 23, Difficulty 5}
\end{subfigure}
\end{figure}

\newpage
\section{Environment Scenarios}
\label{appendix:env-scenarios-zoom}
\begin{figure}[H]
\begin{subfigure}{0.495\textwidth}
\centering
\includegraphics[width=\textwidth]{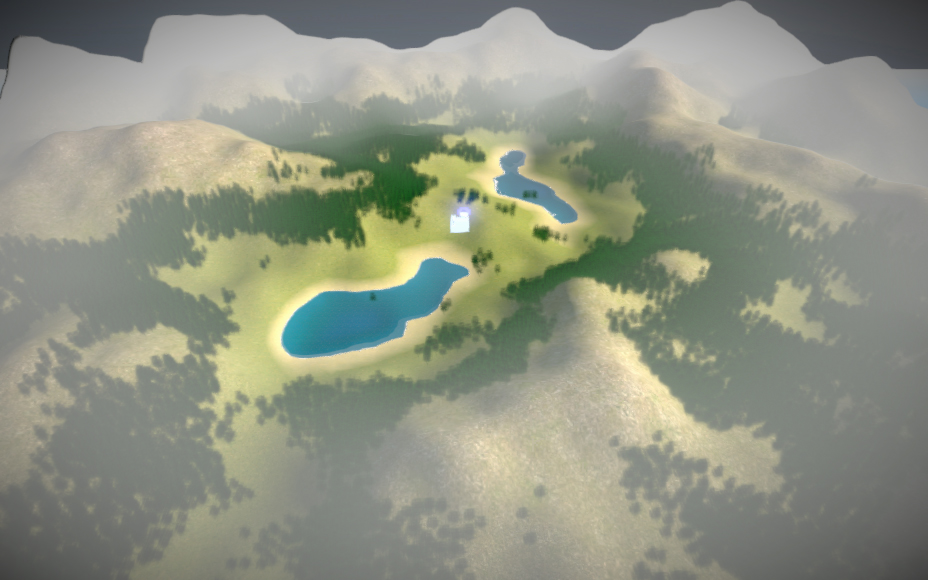}
\caption{Seed 23, Difficulty 5}
\end{subfigure}
\begin{subfigure}{0.495\textwidth}
\centering
\includegraphics[width=\textwidth]{source/seed_25.jpg}
\caption{Seed 25, Difficulty 5}
\end{subfigure}
\begin{subfigure}{0.495\textwidth}
\centering
\includegraphics[width=\textwidth]{source/seed_26.jpg}
\caption{Seed 26, Difficulty 5}
\end{subfigure}
\begin{subfigure}{0.495\textwidth}
\centering
\includegraphics[width=\textwidth]{source/seed_27.jpg}
\caption{Seed 27, Difficulty 5}
\end{subfigure}
\begin{subfigure}{0.495\textwidth}
\centering
\includegraphics[width=\textwidth]{source/seed_28.jpg}
\caption{Seed 28, Difficulty 5}
\end{subfigure}
\begin{subfigure}{0.495\textwidth}
\centering
\includegraphics[width=\textwidth]{source/seed_31.jpg}
\caption{Seed 31, Difficulty 5}
\end{subfigure}
\begin{subfigure}{0.495\textwidth}
\centering
\includegraphics[width=\textwidth]{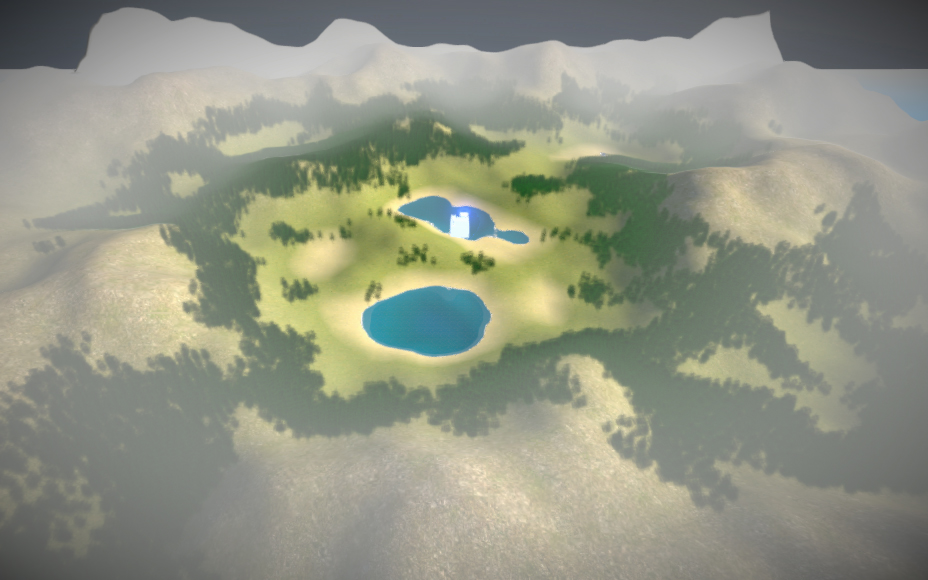}
\caption{Seed 32, Difficulty 5}
\end{subfigure}
\begin{subfigure}{0.495\textwidth}
\centering
\includegraphics[width=\textwidth]{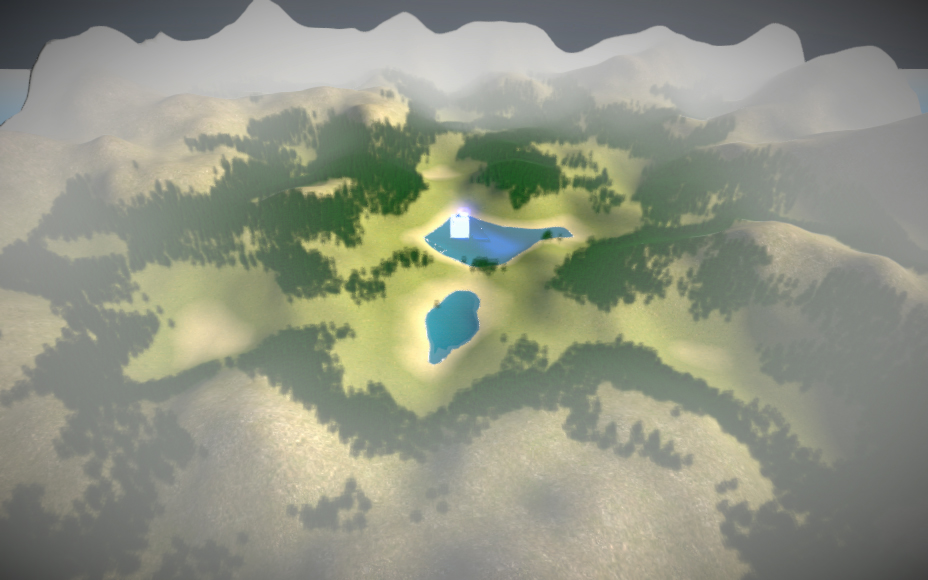}
\caption{Seed 33, Difficulty 5}
\end{subfigure}
\end{figure}

\newpage
\section{Environment Difficulty}
\label{appendix:env-difficulty-zoom}
\begin{figure}[H]
\begin{subfigure}{0.495\textwidth}
\centering
\includegraphics[width=\textwidth]{source/diff_1.jpg}
\caption{Difficulty 1}
\end{subfigure}
\begin{subfigure}{0.495\textwidth}
\centering
\includegraphics[width=\textwidth]{source/diff_3.jpg}
\caption{Difficulty 3}
\end{subfigure}
\begin{subfigure}{0.495\textwidth}
\centering
\includegraphics[width=\textwidth]{source/diff_4.jpg}
\caption{Difficulty 4}
\end{subfigure}
\begin{subfigure}{0.495\textwidth}
\centering
\includegraphics[width=\textwidth]{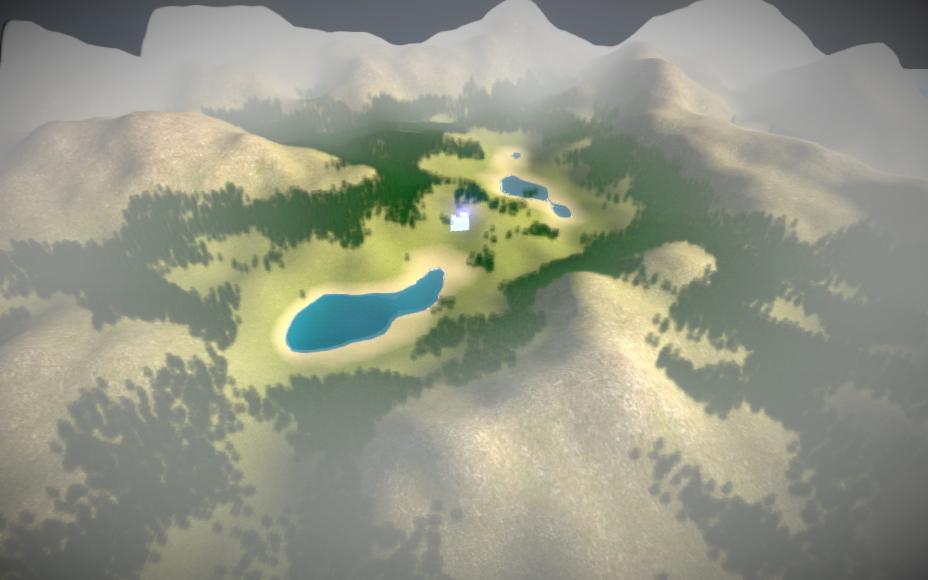}
\caption{Difficulty 6}
\end{subfigure}
\begin{subfigure}{0.495\textwidth}
\centering
\includegraphics[width=\textwidth]{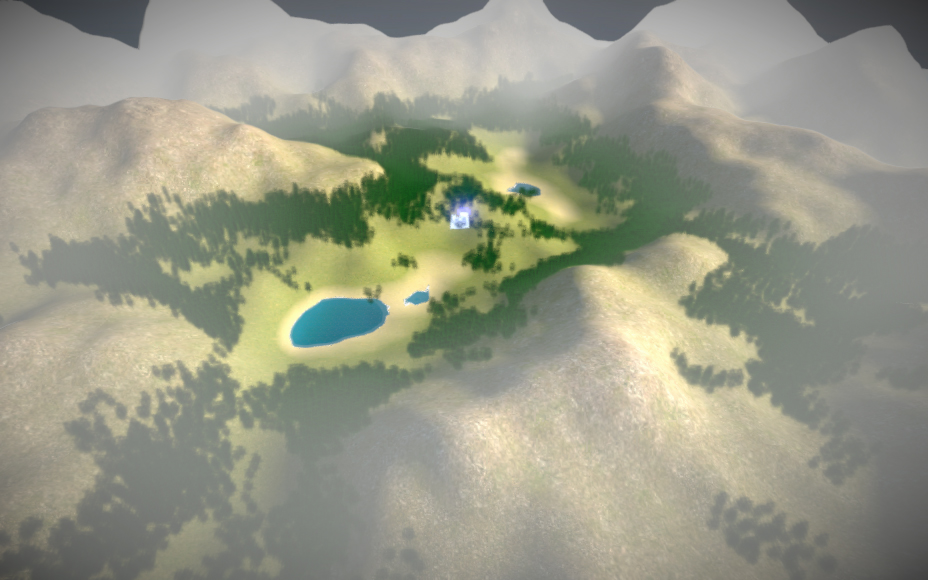}
\caption{Difficulty 7}
\end{subfigure}
\begin{subfigure}{0.495\textwidth}
\centering
\includegraphics[width=\textwidth]{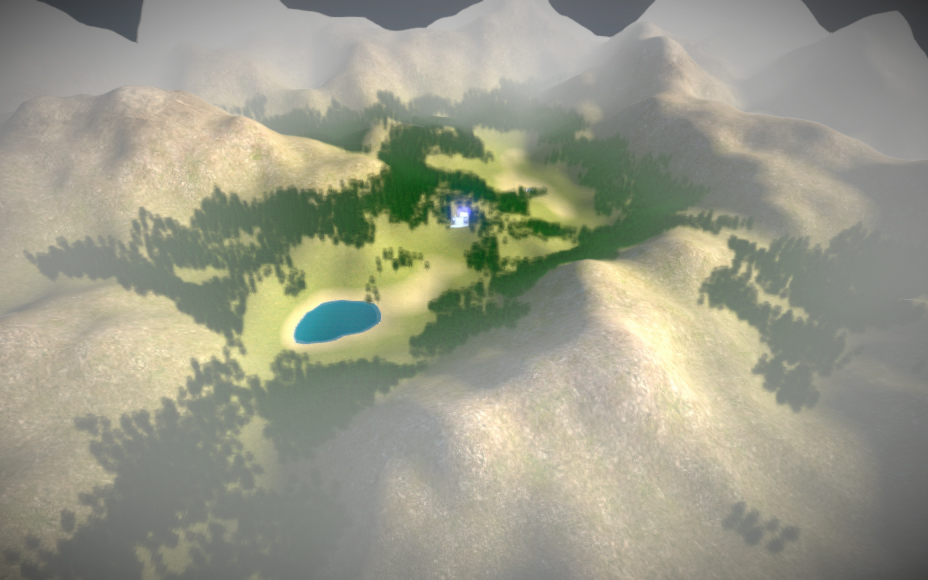}
\caption{Difficulty 8}
\end{subfigure}
\begin{subfigure}{0.495\textwidth}
\centering
\includegraphics[width=\textwidth]{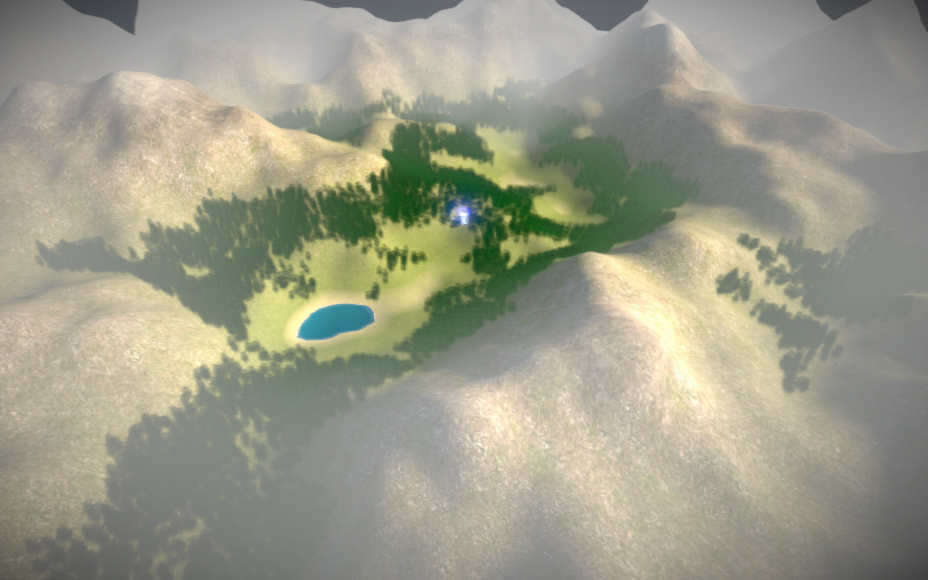}
\caption{Difficulty 9}
\end{subfigure}
\begin{subfigure}{0.495\textwidth}
\centering
\includegraphics[width=\textwidth]{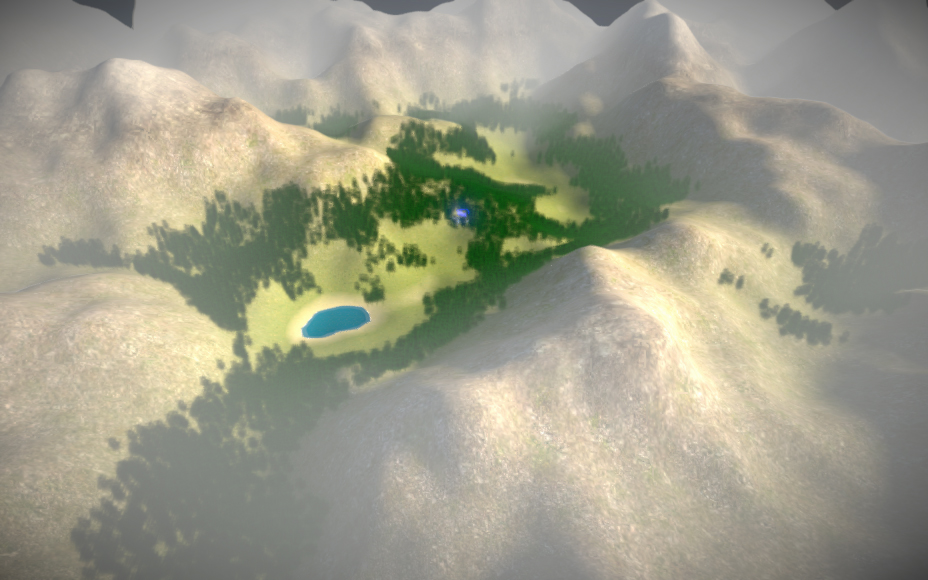}
\caption{Difficulty 10}
\end{subfigure}
\end{figure}

\newpage
\section{Environment Scenario Samples: Difficulty VS. Seed Matrix}
\label{appendix:difficulty-vs-seed-matrix}
\begin{figure}[H]
    \includegraphics[width=\textwidth]{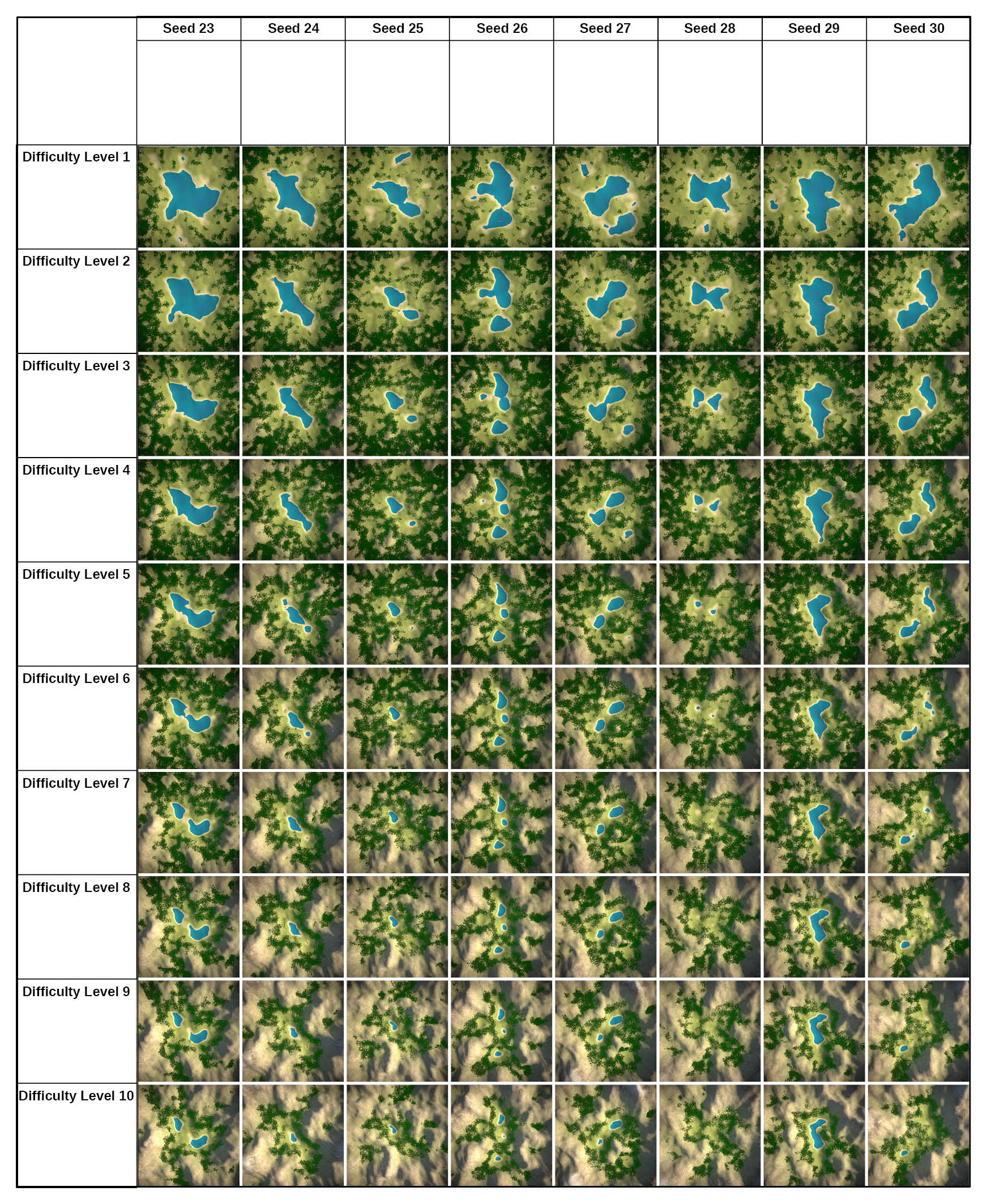}
    \caption{Difficulty Seed Matrix}
\end{figure}

\newpage
\section{Training Data}
\label{appendix:train-results-graphs-zoomed}
\begin{figure}[H]
\begin{subfigure}{0.49\textwidth}
\centering
\includegraphics[width=\textwidth]{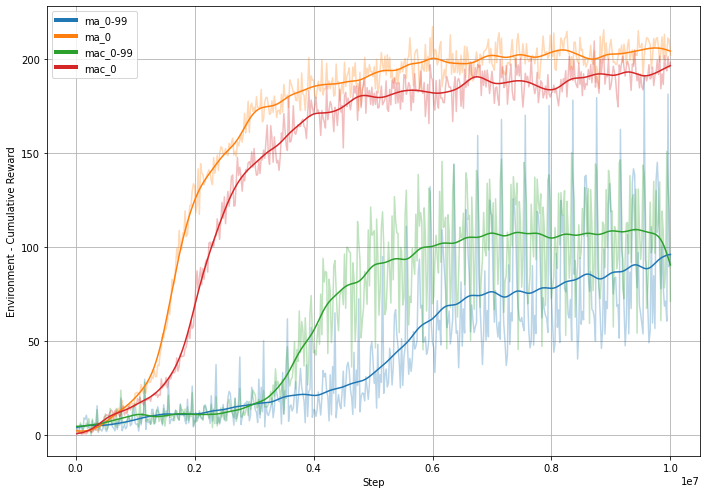}
\caption{Environment - Cumulative Reward}
\end{subfigure}
\begin{subfigure}{0.49\textwidth}
\centering
\includegraphics[width=\textwidth]{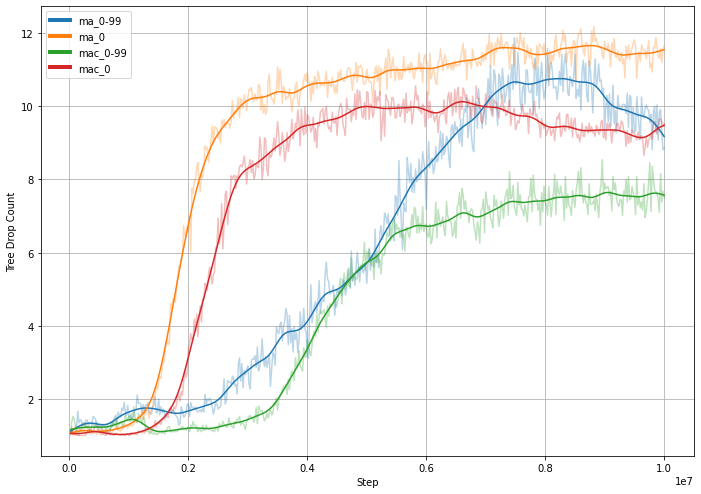}
\caption{Tree Drop Count}
\end{subfigure}
\begin{subfigure}{0.49\textwidth}
\centering
\includegraphics[width=\textwidth]{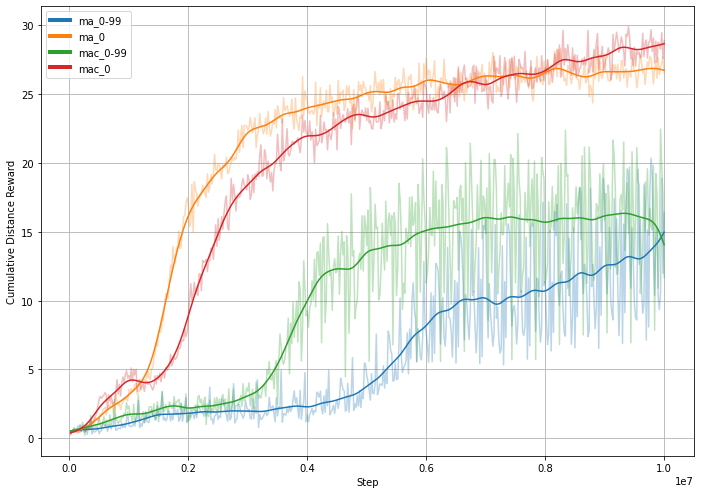}
\caption{Cumulative Distance Reward}
\end{subfigure}
\begin{subfigure}{0.49\textwidth}
\centering
\includegraphics[width=\textwidth]{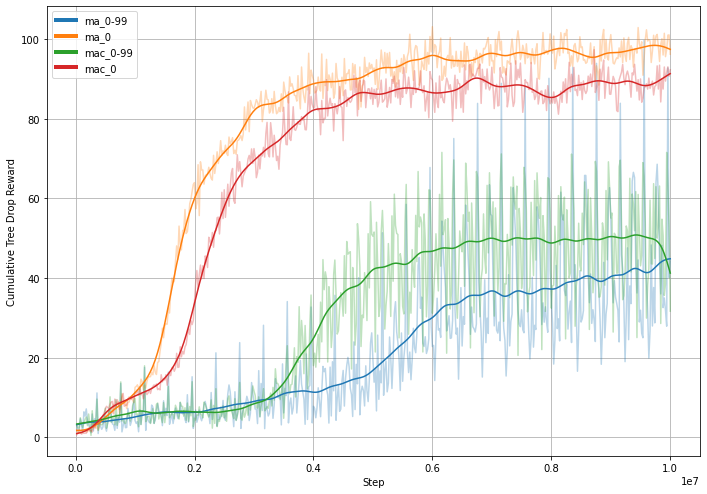}
\caption{Cumulative Tree Drop Reward}
\end{subfigure}
\begin{subfigure}{0.49\textwidth}
\centering
\includegraphics[width=\textwidth]{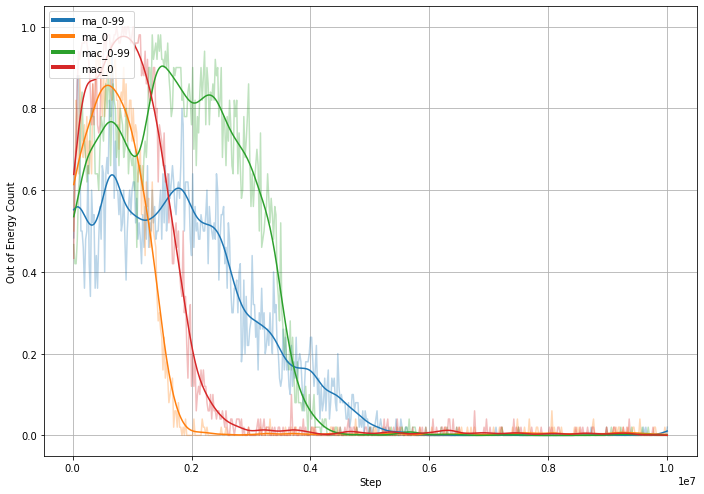}
\caption{Out of Energy Count}
\end{subfigure}
\begin{subfigure}{0.49\textwidth}
\centering
\includegraphics[width=\textwidth]{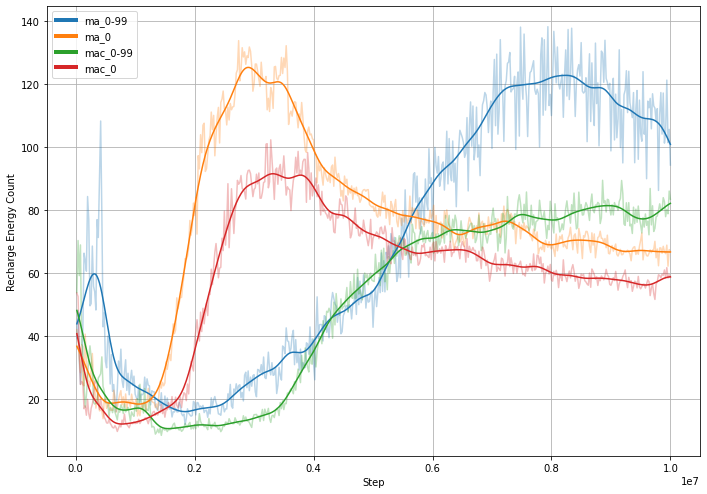}
\caption{Recharge Energy Count}
\end{subfigure}
\begin{subfigure}{0.49\textwidth}
\centering
\includegraphics[width=\textwidth]{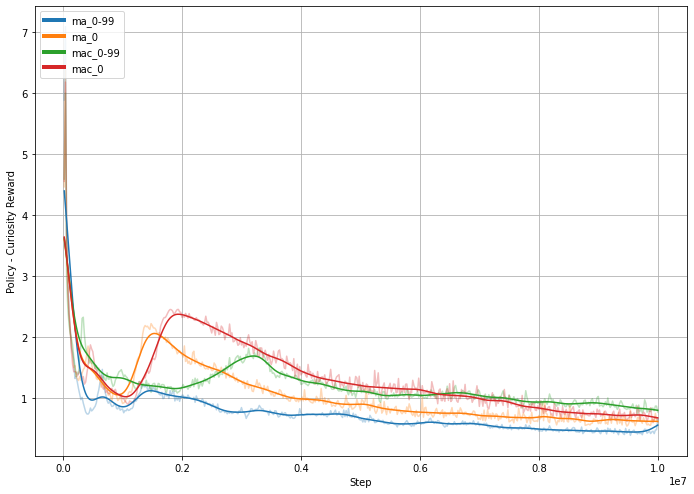}
\caption{Policy - Curiosity Reward}
\end{subfigure}
\begin{subfigure}{0.49\textwidth}
\centering
\includegraphics[width=\textwidth]{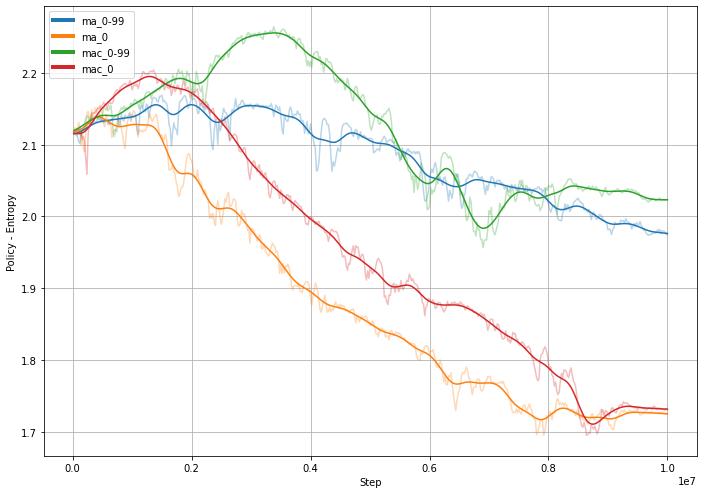}
\caption{Policy - Entropy}
\end{subfigure}
\end{figure}

\section{Test Data: Terrain Scenario Seed 111}
\label{appendix:inference-results-graphs-zoomed}
\begin{figure}[H]
    \begin{subfigure}{0.49\textwidth}
    \centering
    \includegraphics[width=\textwidth]{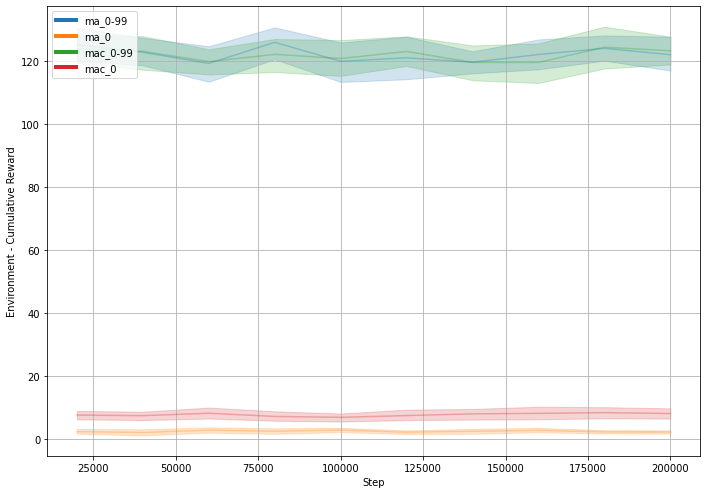}
    \caption{Environment - Cumulative Reward}
    \end{subfigure}
    \begin{subfigure}{0.49\textwidth}
    \centering
    \includegraphics[width=\textwidth]{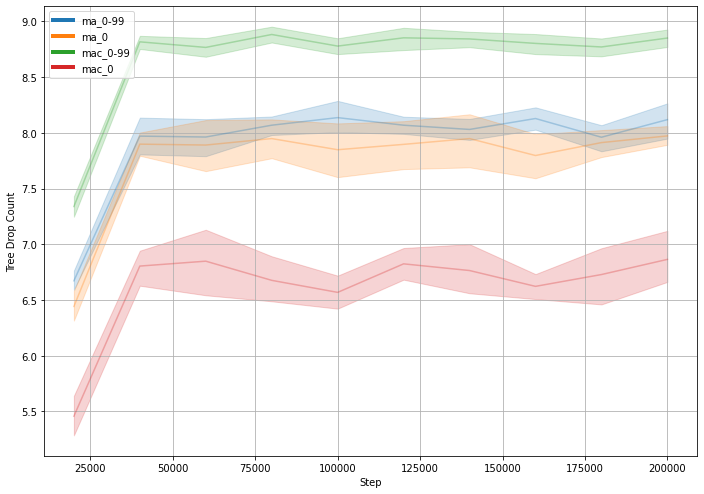}
    \caption{Tree Drop Count}
    \end{subfigure}
    \begin{subfigure}{0.49\textwidth}
    \centering
    \includegraphics[width=\textwidth]{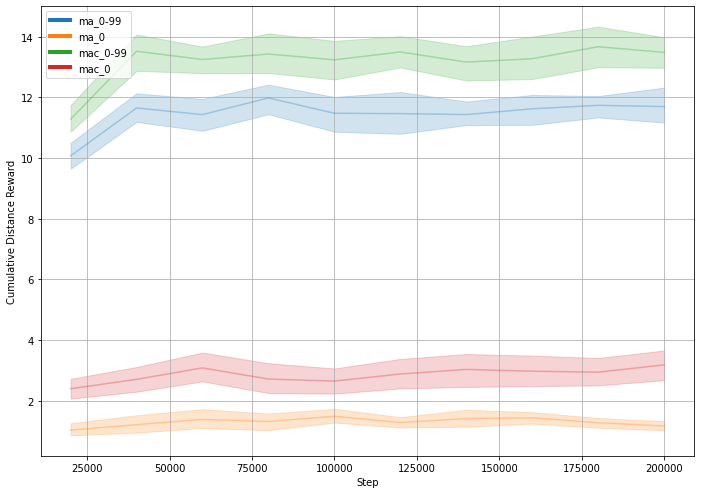}
    \caption{Cumulative Distance Reward}
    \end{subfigure}
    \begin{subfigure}{0.49\textwidth}
    \centering
    \includegraphics[width=\textwidth]{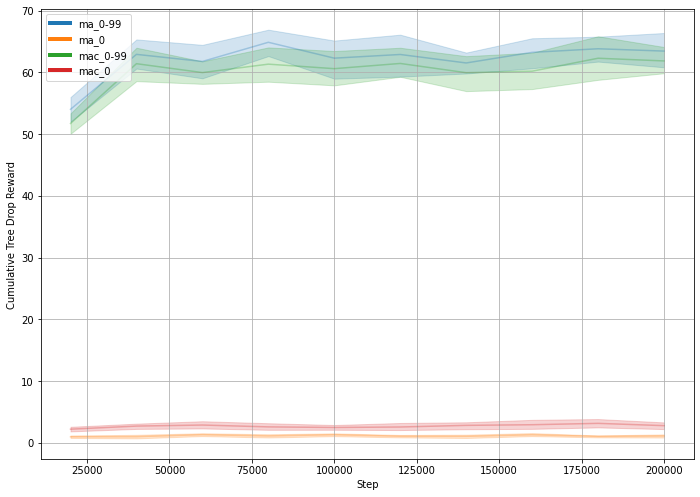}
    \caption{Cumulative Tree Drop Reward}
    \end{subfigure}
    \begin{subfigure}{0.49\textwidth}
    \centering
    \includegraphics[width=\textwidth]{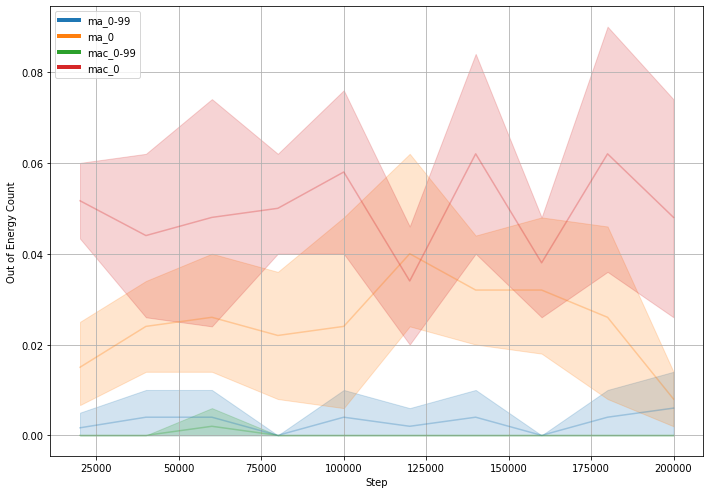}
    \caption{Out of Energy Count}
    \end{subfigure}
    \begin{subfigure}{0.49\textwidth}
    \centering
    \includegraphics[width=\textwidth]{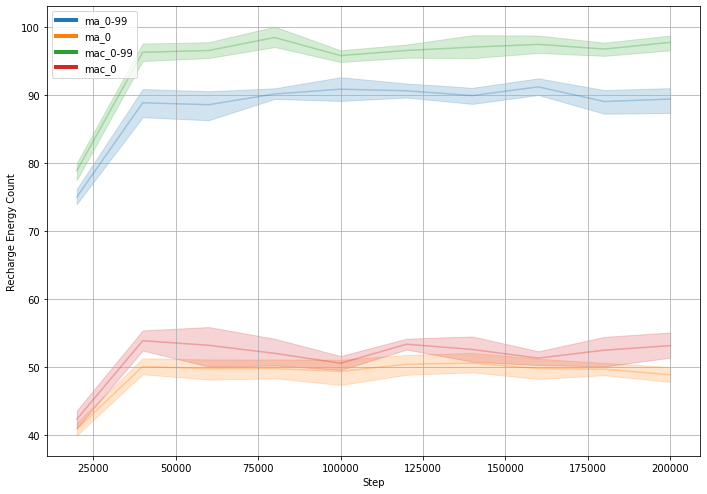}
    \caption{Recharge Energy Count}
    \end{subfigure}
    \begin{subfigure}{0.49\textwidth}
    \centering
    \includegraphics[width=\textwidth]{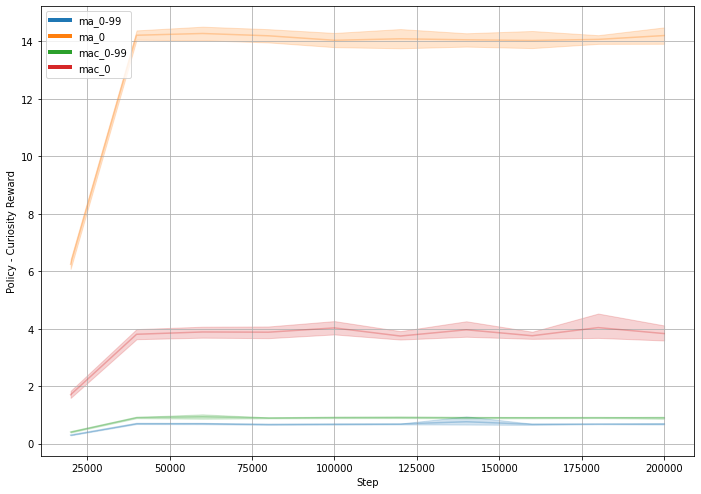}
    \caption{Policy - Curiosity Reward}
    \end{subfigure}
    \begin{subfigure}{0.49\textwidth}
    \centering
    \includegraphics[width=\textwidth]{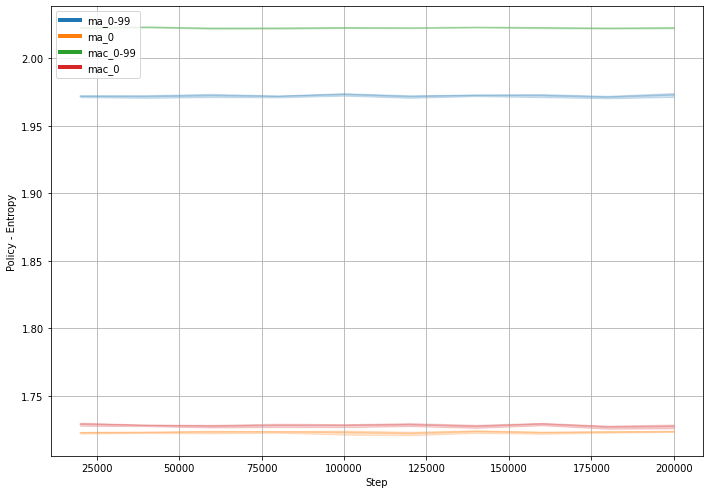}
    \caption{Policy - Entropy}
    \end{subfigure}
\end{figure}

\newpage

\section{Flight Path Plot: Flight Height and Path Sequence}
\label{appendix:Flight-Path-Height-Sequence}
\begin{figure}[H]
    \centering
    \begin{subfigure}{0.47\textwidth}
    \centering
    \includegraphics[width=\textwidth]{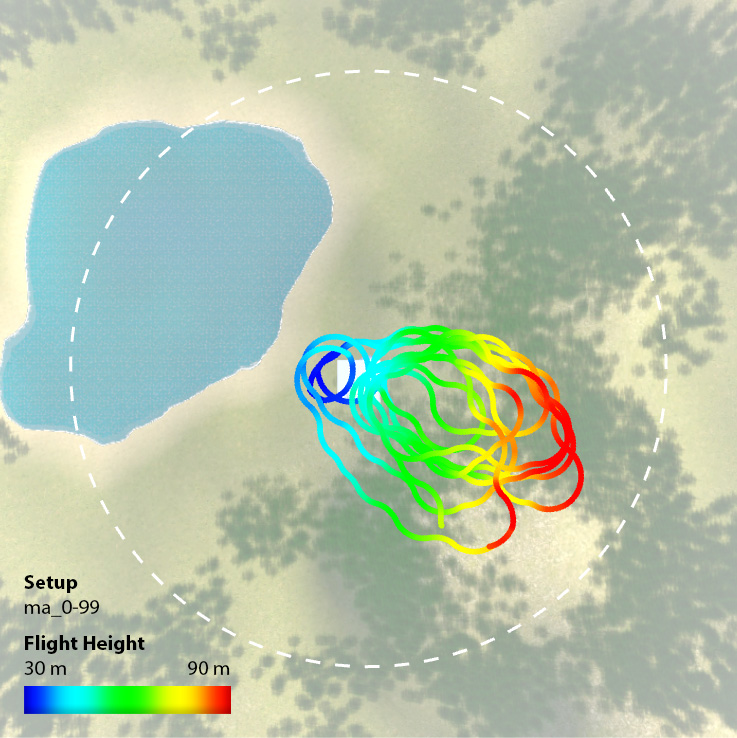}
    \caption{Flight Height: MA 0-99 on Terrain Seed 111}
    \end{subfigure}
    \begin{subfigure}{0.47\textwidth}
    \centering
    \includegraphics[width=\textwidth]{source/flight_path/flight_path__0007_ma_0_99_on_111.jpg}
    \caption{Path Sequence: MA 0-99 on Terrain Seed 111}
    \end{subfigure}
    \begin{subfigure}{0.47\textwidth}
    \centering
    \includegraphics[width=\textwidth]{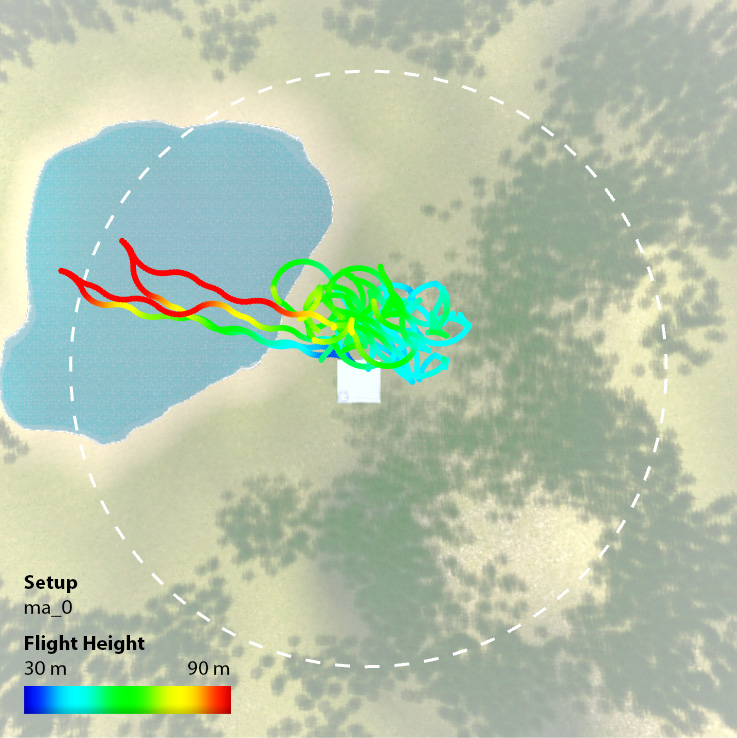}
    \caption{Flight Height: MA 0 on Terrain Seed 111}
    \end{subfigure}
    \begin{subfigure}{0.47\textwidth}
    \centering
    \includegraphics[width=\textwidth]{source/flight_path/flight_path__0006_ma_0_on_111.jpg}
    \caption{Path Sequence: MA 0 on Terrain Seed 111}
    \end{subfigure}
    \begin{subfigure}{0.47\textwidth}
    \centering
    \includegraphics[width=\textwidth]{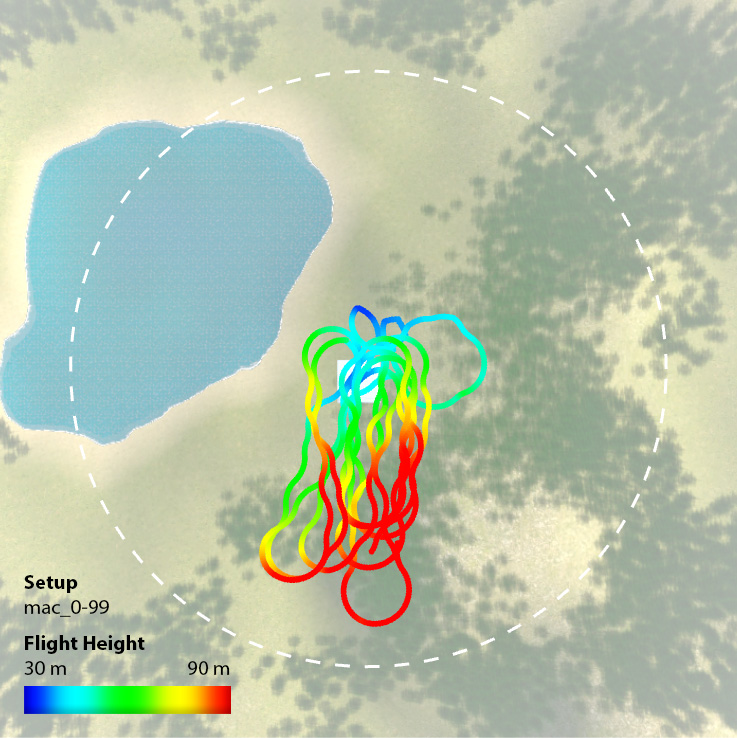}
    \caption{Flight Height: MAC 0-99 on Terrain Seed 111}
    \end{subfigure}
    \begin{subfigure}{0.47\textwidth}
    \centering
    \includegraphics[width=\textwidth]{source/flight_path/flight_path__0005_mac_0_99_on_111.jpg}
    \caption{Path Sequence: MAC 0-99 on Terrain Seed 111}
    \end{subfigure}
\end{figure}

\newpage

\begin{figure}[H]
    \centering
    \begin{subfigure}{0.47\textwidth}
    \centering
    \includegraphics[width=\textwidth]{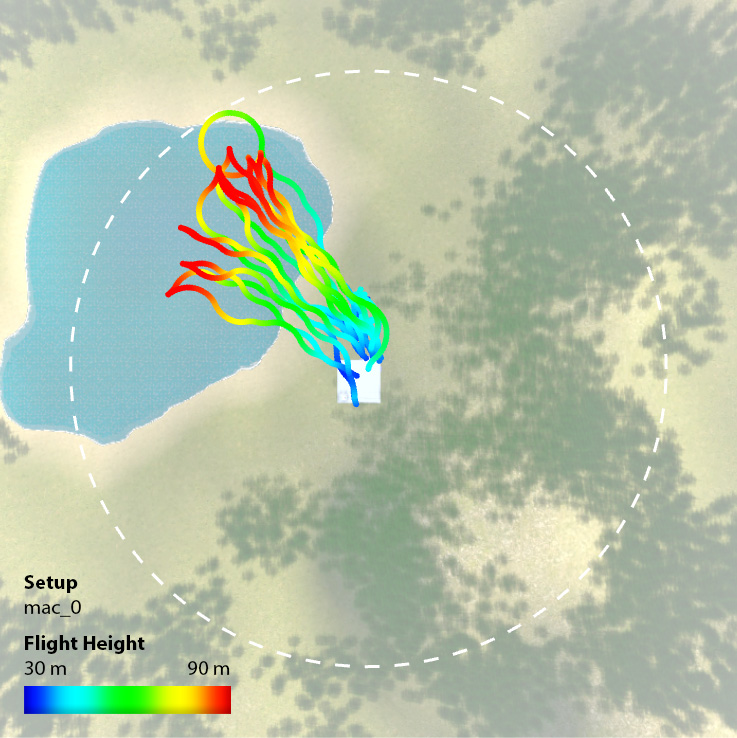}
    \caption{Flight Height: MAC 0 on Terrain Seed 111}
    \end{subfigure}
    \begin{subfigure}{0.47\textwidth}
    \centering
    \includegraphics[width=\textwidth]{source/flight_path/flight_path__0004_mac_0_on_111_path.jpg}
    \caption{Path Sequence: MAC 0 on Terrain Seed 111}
    \end{subfigure}
\end{figure}

\subsection{2000 Time Step Inference on Terrain Scenario Seed 111: Setup Comparison}
\label{appendix:2000-Step-Comaprison-on-111}
\begin{figure}[H]
    \centering
    \begin{subfigure}{0.94\textwidth}
    \centering
    \includegraphics[width=\textwidth]{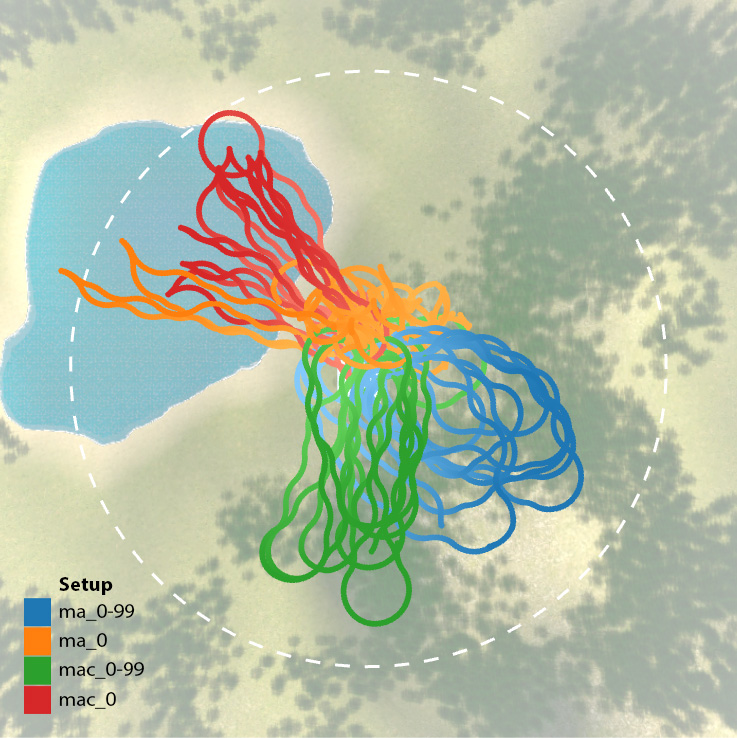}
    \caption{MA 0-99, MA 0, MAC 0-99, MAC 0 - 2000 Steps Inference}
    \end{subfigure}
\end{figure}

\subsection{1e6 Time Step Inference on Terrain Scenario Seed 111}
\label{appendix:1e6-Step-Inference-on-111}
\begin{figure}[H]
    \centering
    \begin{subfigure}{0.47\textwidth}
    \centering
    \includegraphics[width=\textwidth]{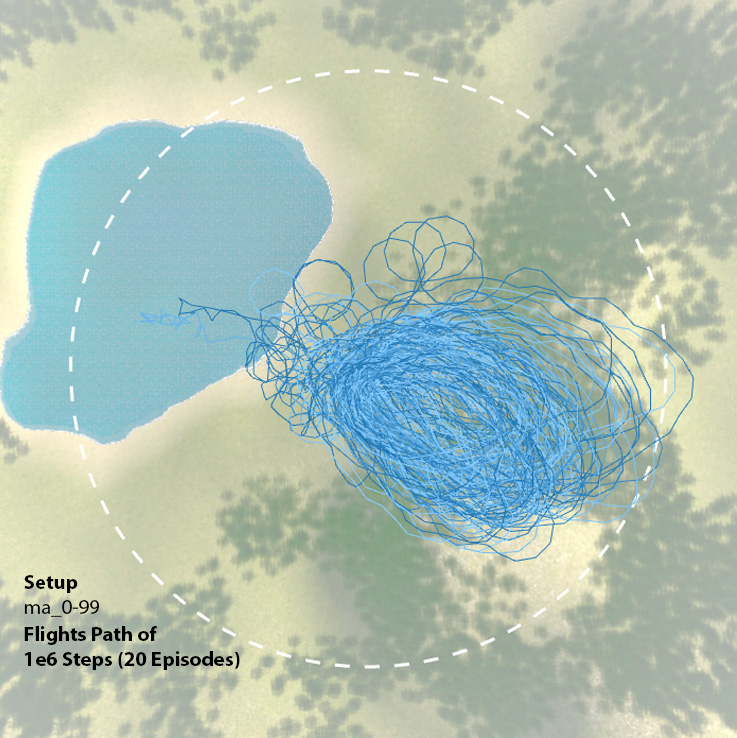}
    \caption{MA 0-99 - 1e6 Steps Inference}
    \end{subfigure}
    \begin{subfigure}{0.47\textwidth}
    \centering
    \includegraphics[width=\textwidth]{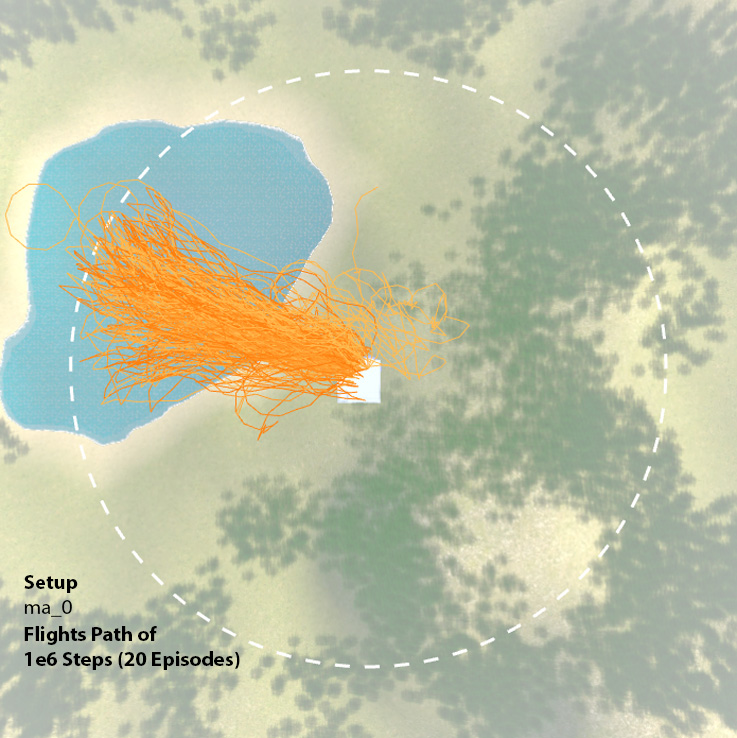}
    \caption{MA 0 - 1e6 Steps Inference}
    \end{subfigure}
    \\
    \begin{subfigure}{0.47\textwidth}
    \centering
    \includegraphics[width=\textwidth]{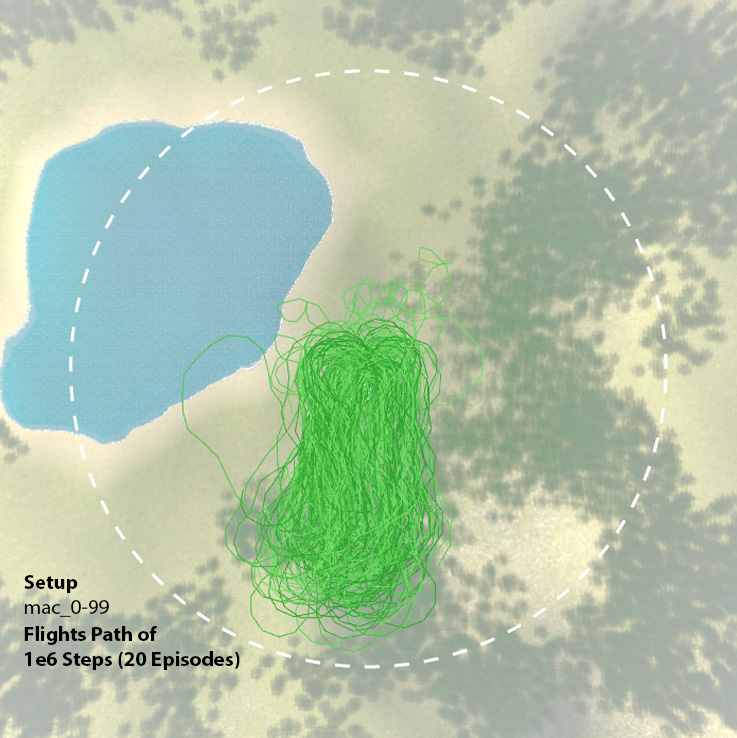}
    \caption{MAC 0-99 - 1e6 Steps Inference}
    \end{subfigure}
    \begin{subfigure}{0.47\textwidth}
    \centering
    \includegraphics[width=\textwidth]{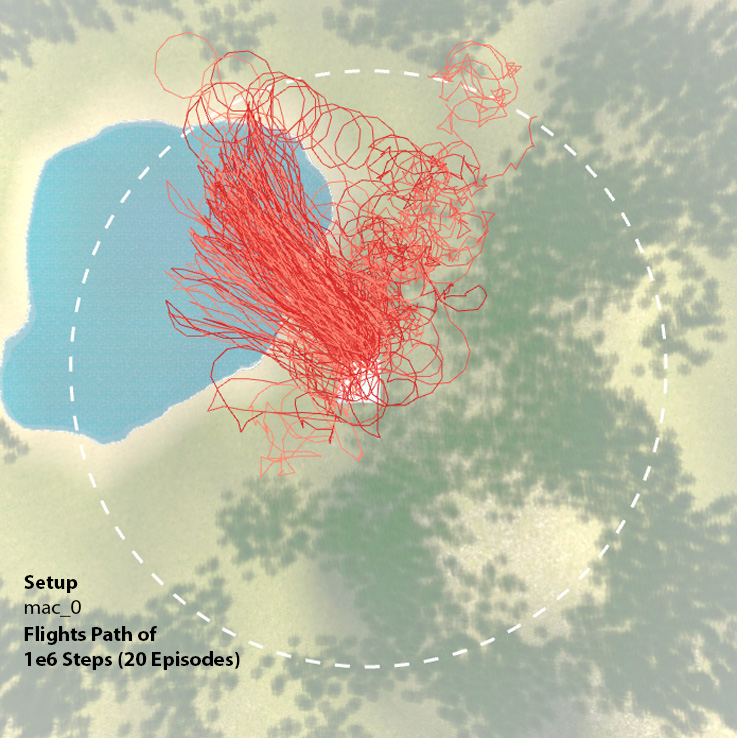}
    \caption{MAC 0 - 1e6 Steps Inference}
    \end{subfigure}
\end{figure}

\end{document}